\documentclass{article}

\PassOptionsToPackage{table}{xcolor}
\PassOptionsToPackage{numbers}{natbib}
\usepackage[eandd, preprint]{neurips_2026}

\usepackage[most]{tcolorbox}
\tcbuselibrary{breakable}
\usepackage[utf8]{inputenc} 
\usepackage[T1]{fontenc}    
\usepackage{hyperref}       
\usepackage{url}            
\usepackage{booktabs}       
\usepackage{amsfonts}       
\usepackage{amsmath}
\usepackage{amssymb}       
\usepackage{nicefrac}       
\usepackage{microtype}      
\usepackage{xcolor}         
\usepackage{makecell}
\usepackage{enumitem}

\usepackage{threeparttable}
\usepackage{adjustbox}
\usepackage{multirow}
\usepackage{makecell}
\usepackage{enumitem}
\usepackage{subcaption}
\usepackage{soul}
\usepackage{pifont}
\usepackage{utfsym}
\usepackage{bbding}
\usepackage{indentfirst}
\usepackage{algorithm}
\usepackage{algorithmic}
\usepackage{multicol}
\usepackage{soul}
\usepackage{xcolor}
\definecolor{lightskyblue}{RGB}{220,238,255}
\definecolor{bestblue}{RGB}{220,238,255}
\definecolor{secondblue}{RGB}{238,245,255}
\sethlcolor{lightskyblue}  
\newcommand{\rowhl}{\rowcolor{lightskyblue}} 

\usepackage{pdflscape}
\usepackage{rotating}
\usepackage{afterpage} 
\usepackage{threeparttable}
\usepackage[table]{xcolor}
\definecolor{lightgreen}{HTML}{D0EDD0}
\definecolor{promptblue}{RGB}{38, 77, 115}
\newtcolorbox{promptbox}[2][]{
    enhanced,
    colback=white,                 
    colframe=black!75,             
    boxrule=0.8pt,                 
    arc=3mm,                       
    auto outer arc,                
    fonttitle=\bfseries,           
    colbacktitle=promptblue,       
    coltitle=white,                
    breakable,                     
    title={#2},                  
    #1                             
}
\newcommand{\cmark}{\checkmark}
\newcommand{\xmark}{$\times$}
\title{Conditional Multi-Event Temporal Grounding in Long-Form Video}

\author{%
  \normalfont
  Yuanhao Zou\textsuperscript{1} \quad
  Arthad Kulkarni\textsuperscript{1} \quad
  Lucas To\~nanez\textsuperscript{1} \quad
  Lincoln Spencer\textsuperscript{1} \quad
  Guangyu Sun\textsuperscript{1} \\
  Tianxingjian Ding\textsuperscript{1} \quad
  Andong Deng\textsuperscript{1} \quad
  Yi Li\textsuperscript{2} \quad
  Shuangjun Liu\textsuperscript{2} \quad
  Yuan Li\textsuperscript{2} \\
  Dashan Gao\textsuperscript{2} \quad
  Ning Bi\textsuperscript{2} \quad
  Taotao Jing\textsuperscript{2} \quad
  Shuai Zhang\textsuperscript{2} \quad
  Chen Chen\textsuperscript{1,}\thanks{Corresponding author. \texttt{chen.chen@ucf.edu}} \\[6pt]
  \textsuperscript{1}University of Central Florida \qquad
  \textsuperscript{2}Qualcomm AI Research
}

\begin{document}

\maketitle

\begin{abstract}
Multimodal large language models have made rapid progress in video temporal grounding, yet real-world applications routinely require localizing \emph{every} event that satisfies compositional temporal and spatial conditions. Existing benchmarks fall short: they localize only a single moment per query, count without temporal conditions, or treat grounding and counting as disjoint tasks. We introduce \textbf{CoMET-Bench} for Conditional Multi-Event Temporal Grounding in long-form video, comprising 2{,}789 queries over 600 videos averaging 33.8 minutes across five real-world domains, with each query composed from 4 temporal conditions, 3 spatial conditions, and a dedicated negative-query subset. We further propose a unified evaluation protocol jointly measuring counting, grounding, and negative-query recognition, including a new \textbf{Rejection-F1} metric that prevents trivial gaming by lazy ``always-empty'' models. Benchmarking a broad suite of MLLMs, agent-based, and grounding-specialized methods reveals that existing approaches remain far from solving this task. Building on these findings, we propose \textbf{CoMET-Agent}, a training-free agentic framework that reformulates the task as structured search-and-aggregate, improving F1@0.5 by 6.1\% over GPT-5 purely through structural reasoning. Failure analysis further surfaces three open directions: fine-grained entity tracking, position-uniform retrieval, and causal event pairing.
Project page: \url{https://alexco1d.github.io/comet/}.
\end{abstract}

\section{Introduction}
\label{introduction}

Multimodal large language models (MLLMs)~\cite{gpt5, google_gemini3, bai2025qwen3vl, wang2025internvl3_5} have reshaped the paradigm of video understanding by enabling flexible, query-driven reasoning over long and complex videos. Beyond recognizing visual content, modern MLLMs can answer questions~\cite{data-egoschema, data-videomme, data-wu2024longvideobench, data-zhou2025mlvu}, generate descriptions~\cite{chen2023vast, activitynet-captions}, and ground moments~\cite{gao2017tall-charades-sta-data, lei2020tvr-data, qvhighlight, jiang2014thumos-data} in response to natural-language queries, showing strong promise on a wide range of real-world tasks. Among these capabilities, \emph{Video Temporal Grounding} (VTG), which aims to localize the time interval in a video that corresponds to a natural-language query, has emerged as a fundamental building block for downstream applications, with rapid progress on benchmarks~\cite{yuan2025momentseeker, chandrasegaran2024hourvideo, liu2024etbench} and dedicated methods~\cite{ren2024timechat, huang2024vtimellm, guo2024trace, wang2024groundedvideollm}.

\begin{figure}[t]
  \centering
  \includegraphics[width=1\linewidth]{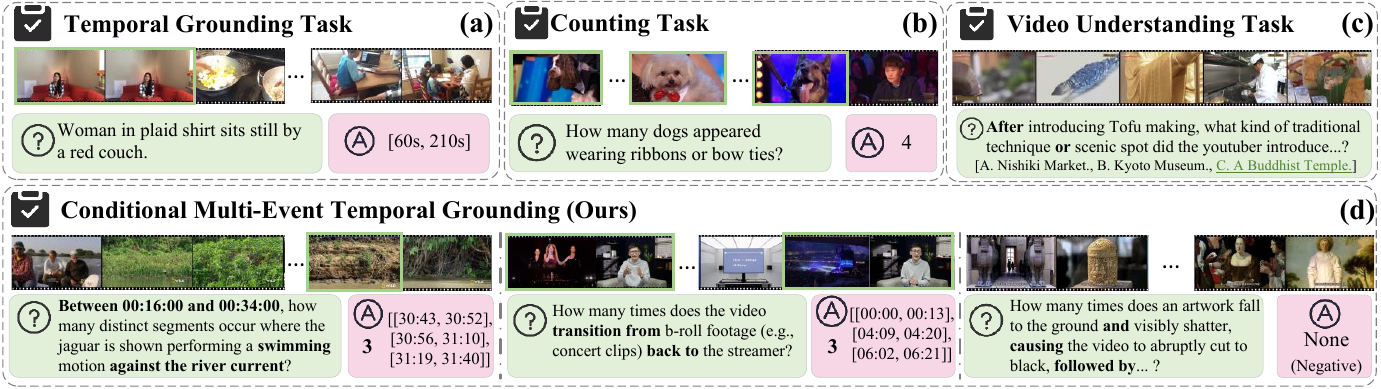}
  \caption{\textbf{Conditional Multi-Event Temporal Grounding compared with prior video tasks.} \textbf{(a)} \emph{Temporal Grounding}  pairs a query with a single moment. \textbf{(b)} \emph{Counting}  returns a single integer without temporal evidence. \textbf{(c)} \emph{Video Understanding}  selects an answer from predefined choices under a temporal condition. \textbf{(d)} Our task requires the model to (i)~parse a compositional query that may include temporal and spatial conditions, (ii)~ground \emph{every} qualifying event with precise timestamps, and (iii)~return timestamps with an exact count, while allowing an empty set for negative queries.}
  \label{fig:intro}
  \vspace{-8pt}
\end{figure}

\begin{figure}[t]
  \centering
  \includegraphics[width=1\linewidth]{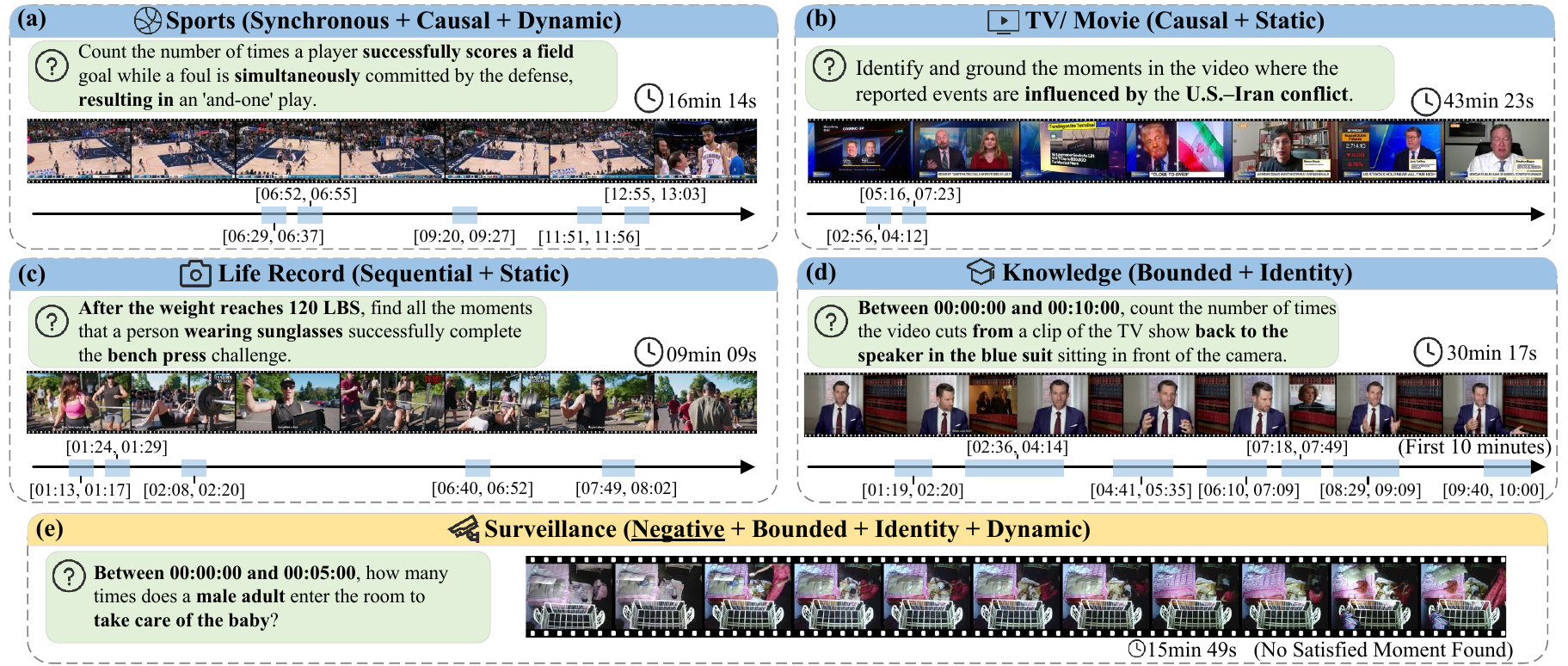}
  \caption{\textbf{Example queries from CoMET-Bench across the five video domains.} Each panel shows a representative query, its compositional attribute combination drawn from four temporal conditions (Causal, Sequential, Synchronous, Bounded) and three spatial conditions (Static, Dynamic, Identity), the source video duration, and all ground-truth grounded intervals on the timeline. The benchmark spans \emph{Sports}, \emph{TV \& Movie}, \emph{Life Record}, \emph{Knowledge}, and \emph{Surveillance} (highlighted). The surveillance example is a \emph{Negative} query whose temporal condition yields zero matching instances, evaluating the model's ability to resist hallucinating nonexistent events.}
  \label{fig:benchmark detail}
  \vspace{-6pt}
\end{figure}

However, in many real-world applications, such as video editing and sports analysis, temporal grounding often requires more than identifying a \emph{single} relevant moment. A video editor may ask: \emph{``How many times does the video transition from b-roll footage back to the streamer?''} (Fig.~\ref{fig:intro} (d)); a basketball analyst may ask: \emph{``How many times does a player successfully score a field goal while a foul is simultaneously committed by the defense, resulting in an `and-one' play?''} (Fig.~\ref{fig:benchmark detail} (a)). Such queries are \textbf{conditional, time-bounded, and correspond to multi-event}: a model must search within a specified scope, identify every distinct occurrence that satisfies the spatial and temporal conditions, and produce an exact count supported by grounded start/end timestamps.



 
Despite recent progress, no existing benchmark fully evaluates this capability along three dimensions.
\textbf{(1)~Temporal output.} The majority of VTG benchmarks~\cite{gao2017tall-charades-sta-data, activitynet-captions, qvhighlight, yuan2025momentseeker} are designed to localize a \emph{single} moment per query and do not require the model to retrieve \emph{multiple} event instances~\cite{jin2026TG-paradigms}. Conversely, counting-oriented benchmarks~\cite{hu2022transrac-repcount-data, ucfrep-data, countix-data} count repetitive physical actions (e.g., push-ups, squats) instead of semantically diverse events.
\textbf{(2)~Query complexity.} Existing counting benchmarks~\cite{hu2022transrac-repcount-data, ucfrep-data, countix-data, tsuchiya2026ec-bench} are largely \emph{object-dominated}: identifying the target object or repetitive pattern is often sufficient to ground the right moments or answer questions, since their queries impose no temporal condition (e.g., ``after event~Y, events~X and Z occur synchronously'') that would restrict the counting scope. While some video understanding benchmarks~\cite{data-zhou2025mlvu, data-videomme,  nie2024slowfocus, cai2024temporalbench} do involve those temporal conditional queries, they only require the model to select an answer from predefined choices rather than explicitly ground every qualifying event.
\textbf{(3)~Evaluation protocol.} Existing methods may perform well on single-moment grounding and yet fail entirely when it must ground \emph{all} instances of an event under a temporal condition and report their count.
Furthermore, although a recent benchmark~\cite{moment_of_untruth_negative} introduces \emph{negative queries}, whose answer is the empty set, to expose hallucination behavior, it remains confined to single-moment retrieval and unconditioned queries. As summarized in Tab.~\ref{tab:benchmark_comparison}, no benchmark to date mimics the real-world scenario that unifies multi-event grounding, temporal conditions, and negative queries within an evaluation framework.
 
To bridge this gap, we introduce \textsc{\textbf{CoMET-Bench}} for \textbf{\underline{Co}nditional \underline{M}ulti-\underline{E}vent \underline{T}emporal Grounding} in long-form video. CoMET-Bench contains 2{,}789 queries over 600 videos averaging 33.8 minutes, spanning five real-world domains (Sports, TV \& Movie, Life Record, Knowledge, and Surveillance) to capture the diversity of practical scenarios. The benchmark is designed to address the three limitation dimensions identified above. (1) To overcome the \emph{single-moment} restriction, every annotation provides the grounded timestamps of \emph{all} qualifying event instances, requiring exhaustive multi-event instance retrieval rather than a single best moment. (2) To overcome the \emph{unconditional counting} restriction, each query is composed from four temporal conditions (\emph{causal}, \emph{sequential}, \emph{synchronous}, and \emph{bounded}) and three spatial conditions (\emph{static}, \emph{dynamic}, and \emph{identity}), yielding a compositional query space that systematically tests reasoning under diverse temporal scopes. (3) To overcome the \emph{disjoint evaluation} restriction, CoMET-Bench unifies three complementary aspects of evaluation under a single protocol: \textbf{multi-event counting} (set-level) and \textbf{temporal event grounding} (instance-level) for positive queries, and \textbf{negative query recognition} for the dedicated subset of \emph{negative queries} that have zero matching instances. For the last aspect, we further propose \emph{Rejection-F1}, a new metric defined as the harmonic mean of negative-rejection rate and positive coverage, which prevents the trivial ``always-empty'' strategy that can game the standard rejection rate. Together, these designs make CoMET-Bench a rigorous litmus test of whether a model has constructed a genuine temporal representation of the video or is merely performing shallow pattern matching.

We further propose \textsc{\textbf{CoMET-Agent}}, a training-free agentic framework that models long-form video as a \textbf{video graph} and formulates this task as a structured \textbf{search-and-aggregate} problem. CoMET-Agent comprises three components: (i)~a hierarchical \emph{Video Temporal Graph} that organizes the video into a set of searchable event nodes and action nodes, addressing the need for exhaustive search under compositional temporal conditions; (ii)~an MLLM agent that decomposes the query and performs iterative traversal over the graph, addressing the need for query parsing and fine-grained verification; and (iii)~a persistent \emph{Global Memory Bank} that accumulates verified instances with their temporal indices, addressing the need for identity-aware global perception across the full video.
 
Our main contributions are summarized as follows:
\begin{list}{$\bullet$}{%
    \setlength{\leftmargin}{1.2em}%
    \setlength{\itemsep}{0pt}%
    \setlength{\topsep}{0pt}%
    \setlength{\parsep}{0pt}%
    \setlength{\parskip}{0pt}%
  }
\item We introduce \textbf{CoMET-Bench}, the first benchmark for conditional multi-event temporal grounding in long-form video, together with a unified evaluation protocol and a new \textbf{Rejection-F1} metric that jointly measures counting, grounding, and negative-query recognition.
\item We propose \textbf{CoMET-Agent}, a training-free agentic framework that solves this task through structured search-and-aggregate over a Video Temporal Graph with persistent global memory.
\item Extensive benchmarking shows that existing methods consistently fail on this task while CoMET-Agent yields substantial gains, and our analysis surfaces concrete directions for future work.
\end{list}

\begin{table}[t]
  \centering
  \small
  \setlength{\tabcolsep}{2pt}
  \caption{\textbf{Comparison with various video benchmarks.} Answer Type: multiple-choice questions (MC), numeric answers (NA), timestamps (TS), pixel masks (Masks), clip alignment (Clips), or open-ended answers (OE). A dash indicates a task not expressed as conditional event enumeration or a statistic not comparable across the original source corpora. SynopGround maps each paragraph to one interval, rather than all occurrences of a single query. Three key dimensions differentiate our work: \textbf{Multi-Event}: whether queries require grounding multi-event when answering; \textbf{Temp.~Cond.}: whether queries attach temporal conditions that restrict the grounding scope; \textbf{Neg.~Query}: whether the benchmark includes queries with zero matching instances.}
  \label{tab:benchmark_comparison}
  \small
  \renewcommand{\arraystretch}{0.75}
  \begin{tabular}{lcccccccc}
  
    \toprule
    \textbf{Benchmark} & \textbf{Label} & \textbf{\#Videos} & \makecell{\textbf{Average} \\ \textbf{Dur. (s)}} & \textbf{\#Queries} & \makecell{\textbf{Answer}\\\textbf{Type}} & \makecell{\textbf{Multi-}\\\textbf{Event}} & \makecell{\textbf{Temp.}\\\textbf{Cond.}} & \makecell{\textbf{Neg.}\\\textbf{Query}} \\
    \midrule
    \multicolumn{9}{l}{\color{gray}\textit{General video understanding benchmarks}} \\
    Video-MME~\cite{data-videomme} & Human & 900 & 1{,}021.3 & 2{,}700 & MC & \xmark & \cmark & \xmark \\
    MLVU~\cite{data-zhou2025mlvu} & Human & 349 & 905.8 & 502 & MC & \cmark & \cmark & \xmark \\
    LongVideoBench~\cite{data-wu2024longvideobench} & Human & 753 & 574.9 & 1337 & MC & \xmark & \cmark & \xmark \\
    E.T. Bench~\cite{liu2024etbench} & Human & 7002 & 129 & 7289 & OE/MC & \cmark & \cmark & \xmark \\
    \midrule
    \multicolumn{9}{l}{\color{gray}\textit{Counting benchmarks}} \\
    RepCount~\cite{hu2022transrac-repcount-data} & Human & 1{,}451 & 29.4 & 19{,}280 & NA & \xmark & \xmark & \xmark \\
    OVR~\cite{dwibedi2024ovr-data} & Auto & 72{,}552 & 2.7 & 72{,}552 & NA & \xmark & \xmark & \xmark \\
    Countix \cite{countix-data} & Human & 8757 & 6.1 & 8757 & NA & \xmark & \xmark & \xmark \\
    EC-Bench~\cite{tsuchiya2026ec-bench} & Human & 152 & 3420.0 & 1{,}699 & OE/NA & \cmark & \xmark & \xmark \\
    \midrule
    \multicolumn{9}{l}{\color{gray}\textit{Temporal grounding benchmarks}} \\
    Charades-STA~\cite{gao2017tall-charades-sta-data} & Human & 1{,}334 & 30.6 & 3{,}720 & TS & \xmark & \xmark & \xmark \\
    QVHighlights~\cite{qvhighlight} & Human & 476 & 150.0 & 1{,}542 & TS & \xmark & \xmark & \xmark \\
    THUMOS14~\cite{jiang2014thumos-data} & Human & 216 & 186.4 & 3457 & TS & \xmark & \xmark & \xmark \\
    MomentSeeker~\cite{yuan2025momentseeker} & Human & 268 & 1201.9 & 1{,}800 & TS & \cmark & \xmark & \xmark \\
    NA-VMR (Charades-STA)~\cite{moment_of_untruth_negative}& Auto & 1334 & 30.6 & 5{,}270 & TS & \xmark & \xmark & \cmark \\
    NA-VMR (QVHighlights)~\cite{moment_of_untruth_negative}& Auto & 476 & 150.0 & 3{,}100 & TS & \xmark & \xmark & \cmark \\
    \midrule
    \multicolumn{9}{l}{\color{gray}\textit{Additional benchmarks compared during review}} \\
    HourVideo~\cite{chandrasegaran2024hourvideo} & Human & 500 & 2{,}742 & 12{,}976 & MC & \xmark & \cmark & \xmark \\
    SynopGround~\cite{verified_synopground} & Human & 3{,}987 & 2{,}608.4 & 36{,}002 & TS & \xmark & --- & \xmark \\
    Long-RVOS~\cite{verified_longrvos} & Human & 2{,}193 & 60.3 & 24{,}689 & Masks & \xmark & --- & \xmark \\
    IV-Bench~\cite{verified_ivbench} & Human & 967 & 537 & 2{,}585 & MC & \xmark & \cmark & \xmark \\
    HAVEN~\cite{verified_haven} & --- & --- & --- & --- & Clips & \xmark & --- & \xmark \\
    \midrule
    \rowhl\textbf{CoMET-Bench (Ours)} & Human & 600 & 2028.6 & 2,789 & TS/NA & \cmark & \cmark & \cmark \\
    \bottomrule
  \end{tabular}
  \vspace{-10pt}
\end{table}

\section{Related Works}
\label{sec:related_works}

\noindent\textbf{Temporal Grounding Benchmarks.}
Temporal grounding benchmarks evaluate whether models can localize a described event to a precise temporal interval, and have driven steady progress on short-clip moment retrieval~\cite{gao2017tall-charades-sta-data, activitynet-captions, anne2017didemo-data, lei2020tvr-data, qvhighlight} and temporal action localization with closed action vocabularies~\cite{jiang2014thumos-data, caba2015activitynet}. Newer efforts extend grounding to long-form, egocentric, and movie-length videos with richer query formats~\cite{grauman2022ego4d-data, han2023mad-data, liu2024etbench, cai2024temporalbench, chandrasegaran2024hourvideo}, while MomentSeeker~\cite{yuan2025momentseeker} introduces multi-event retrieval and \cite{moment_of_untruth_negative} adds negative queries to expose hallucination. However, the queries in these benchmarks are largely \emph{object-dominated}: locating the target entity or action is typically sufficient to recover the grounded interval, since the queries impose no compositional temporal condition (e.g., \emph{after}, \emph{while}, or \emph{within}) that would constrain when a match qualifies.

\noindent\textbf{Conditional Video Understanding Benchmarks.}
Another line of work studies \emph{conditional} reasoning, where the answer depends on temporal, causal, or compositional context. Holistic comprehension benchmarks~\cite{data-videomme, data-zhou2025mlvu, data-wu2024longvideobench, data-egoschema} evaluate long videos through multiple-choice or open-ended QA, while temporal-reasoning benchmarks~\cite{data-nextqa, AGQA-data, liu2024tempcompass, shangguan2024tomato} probe ordering, causality, and fine-grained temporal perception. Although their queries may include temporal conditions, evaluation usually reduces to selecting from multiple choices rather than grounding every qualifying event with precise timestamps.

\noindent\textbf{Counting Benchmarks.}
Counting benchmarks measure quantitative ability, mainly over repetitive physical motion: RepCount~\cite{hu2022transrac-repcount-data}, UCF-Rep~\cite{ucfrep-data}, and Countix~\cite{countix-data} count exercise repetitions through periodicity detection, and OVR~\cite{dwibedi2024ovr-data} scales this setting to 72k open-vocabulary videos. More recent work moves beyond periodicity to broader event counting~\cite{nie2024slowfocus}, with EC-Bench~\cite{tsuchiya2026ec-bench} pairing enumeration with evidence spans in ultra-long videos. These benchmarks generally ask unconditional questions of the form ``how many times does X happen?'' and rarely require grounded start/end timestamps for every counted instance.

\noindent\textbf{Comparison with additional benchmarks.}
HourVideo~\cite{chandrasegaran2024hourvideo} uses five-way multiple choice, with timestamps conditioning questions rather than being predicted. SynopGround~\cite{verified_synopground} assigns one interval per synopsis paragraph; Long-RVOS~\cite{verified_longrvos} predicts referring-object masks. IV-Bench~\cite{verified_ivbench} studies image-grounded QA, and HAVEN~\cite{verified_haven} studies hierarchical multimodal alignment and summarization. CoMET-Bench instead requires the unknown-cardinality set of all qualifying occurrences of one conditional query, including the empty set.

\noindent\textbf{Summary.}
As summarized in Tab.~\ref{tab:benchmark_comparison}, prior benchmarks largely study temporal grounding, conditional reasoning, and counting in isolation. CoMET-Bench unifies all three: it requires multi-event grounding under compositional temporal and spatial conditions, reports grounded intervals with exact counts, and includes a negative-query subset to probe hallucination resistance under such conditions.

\section{CoMET-Bench}
\label{benchmark building} 

\subsection{Benchmark Design}
\label{sec:benchmark_design}
\textbf{Task formulation.}
Given a long-form video $V$ and a natural-language query $Q$, the task of conditional multi-event temporal grounding is to produce a set of grounded event intervals
\begin{equation}
\begin{aligned}
\mathcal{P}(V, Q) &= \big\{(s_i, e_i)\big\}_{i=1}^{N},
\end{aligned}
\label{task formulation}
\end{equation}
where each $(s_i, e_i)$ denotes the start and end timestamps of the $i$-th event instance that satisfies the temporal condition specified in $Q$, and $N = |\mathcal{P}(V, Q)|$ is the exact event count. A correct prediction requires: (1)~Set-level completeness: all $N$ ground-truth instances are retrieved; and (2)~Instance-level temporal precision: each retrieved interval tightly overlaps its ground-truth counterpart. Notably, $N$ can be zero for \emph{negative queries}, where no event in the video satisfies the condition.
 
\textbf{Query taxonomy.}
Each query in CoMET-Bench is composed of two semantic components: a \emph{temporal condition} and a \emph{spatial condition}. The \emph{temporal condition} restricts the temporal relationship between the events of interest. We define four attributes for it, covering the majority of temporal relations in natural language: \textbf{(1)~Causal}, where one event occurs as a direct consequence of another; \textbf{(2)~Sequential}, where one event strictly precedes or follows another in time; \textbf{(3)~Synchronous}, where two events happen simultaneously; and \textbf{(4)~Bounded}, where events fall within a bounded temporal window. The \emph{spatial condition} specifies visual attributes that qualifying events must exhibit, and is categorized into three types based on the visual granularity required to verify them: \textbf{(1)~Static}, an attribute verifiable from a single frame (e.g., \emph{``a person in a black suit''}); \textbf{(2)~Dynamic}, an attribute requiring multiple frames within a single event to verify (e.g., \emph{``a player shooting from the free-throw line''}); and \textbf{(3)~Identity}, an attribute that requires tracking a specific entity across the full video to distinguish it from other similar entities (e.g., \emph{``the referee who firstly issued a yellow card''}). In addition, we introduce a \textbf{Negative} attribute, which references plausible events or objects that could conceivably occur in the video but in fact never happen, yielding a ground-truth count of zero. This attribute explicitly tests whether a model can resist hallucinating nonexistent events.

\subsection{Dataset Construction}
\label{data construction}
Our dataset construction pipeline consists of three stages: video collection, query generation and verification, and annotation generation and verification. The full pipeline (\ref{sec:app_pipeline_overview}), MLLM prompts (\ref{sec:app_query_prompt} and~\ref{sec:app_annotation_prompt}), human verification protocol (\ref{sec:app_human_verification}), annotation statistics (\ref{sec:app_quality_stats}) and evaluation prompt (\ref{sec:app_benchmark_eval_prompt}) are provided in Appendix~\ref{sec:app_data_annotation}.

\textbf{Video collection.}
To ensure broad coverage of real-world scenarios, we curate a set of long-form videos spanning five categories: TV/Movie, Life Record, Sports, Knowledge, and Surveillance. The resulting 600 videos are drawn from existing long-video benchmarks \cite{chen2024cg-bench-data, data-videomme, data-zhou2025mlvu} and open video platforms such as YouTube. The copyright and license clarification is detailed in \ref{sec:app_copyright}.

\textbf{Source composition and verification effort.}
The source breakdown is 214 Video-MME videos, 180 MLVU videos, 198 CG-Bench videos, and eight self-collected YouTube videos (Tab.~\ref{tab:source_composition}). Eleven annotators worked over three months; three also served as senior adjudicators. The reported total effort was approximately 791 person-hours (17 minutes per retained query). All 2,789 queries were independently re-annotated, with 93.5\% direct consensus at tIoU $>0.7$; the remaining 182 queries were adjudicated. Final annotations contain 15,022 event intervals. Per-stage effort and source-specific access terms appear in Tab.~\ref{tab:annotation_effort_rebuttal} and Appendix~\ref{sec:app_copyright}.

\begin{table}[ht]
\centering
\small
\caption{\textbf{CoMET-Bench video sources.} Counts sum to 600; source-specific copyright/access conditions remain in force.}
\label{tab:source_composition}
\begin{tabular}{lrr}
\toprule
\textbf{Source} & \textbf{Videos} & \textbf{Share} \\
\midrule
Video-MME & 214 & 35.7\% \\
MLVU & 180 & 30.0\% \\
CG-Bench & 198 & 33.0\% \\
Self-collected YouTube & 8 & 1.3\% \\
\midrule
Total & 600 & 100\% \\
\bottomrule
\end{tabular}
\end{table}

\textbf{Examples of human correction.}
Human edits resolve ambiguity and errors in event cardinality: an ambiguous ``15:30'' query bound was rewritten as ``00:15:30''; a missed bear-attack/flare-gun instance was added at 00:08:34--00:08:55. Two adjacent squirrel-crossing fragments (00:01:58--00:02:06 and 00:02:06--00:02:11) were merged into one continuous event, while an over-long inventory-screen interval was tightened from 00:31:58--00:32:27 to 00:31:58--00:32:00. These corrections prevent false negatives, duplicate counting, and imprecise boundaries.

\textbf{Query generation and human verification.}
For each video, we prompt an MLLM \cite{google_gemini3} to generate candidate queries under attribute-driven prompting. Each query targets a random combination of one to four attributes drawn from the eight query types defined above (\emph{Causal}, \emph{Sequential}, \emph{Synchronous}, \emph{Bounded}, \emph{Static}, \emph{Dynamic}, \emph{Identity}, and \emph{Negative}), ensuring uniform coverage of the compositional query space. The prompt further enforces the event-centric property: each query must target at least one query type with clear start and end boundaries. We explicitly prohibit \emph{spatial multi-instance counting} (e.g., ``how many birds are in the sky''), as such queries reduce to single-frame object detection and are orthogonal to our goal of evaluating long-range temporal reasoning. Generated queries are then verified by trained human annotators, who verify that each query follows the main content of the video, reads naturally from a human perspective, and is free of social bias. Queries failing any of these criteria are either rewritten or discarded before entering the grounding stage.
 
\textbf{Annotation generation and human verification.}
For each verified query, we prompt the MLLM to produce an initial set of grounding annotations. These MLLM-generated annotations substantially reduce annotation cost by providing a strong starting point for human refinement, but they frequently miss instances, hallucinate boundaries, or misalign timestamps. Trained annotators then review each annotation against the source video, correcting boundaries, adding missed instances, and removing spurious ones, until every event instance satisfies the query. Each verified query-annotation pair is further audited by an independent second annotator, who jointly inspects the query, its grounding annotations, and the source video. Disagreements between the two annotators are adjudicated by a senior reviewer, and only QA pairs that pass this final round are included in the benchmark.

\subsection{Dataset Statistics}
\label{data statistics}
\begin{figure*}[t]
  \centering
  \includegraphics[width=1\linewidth]{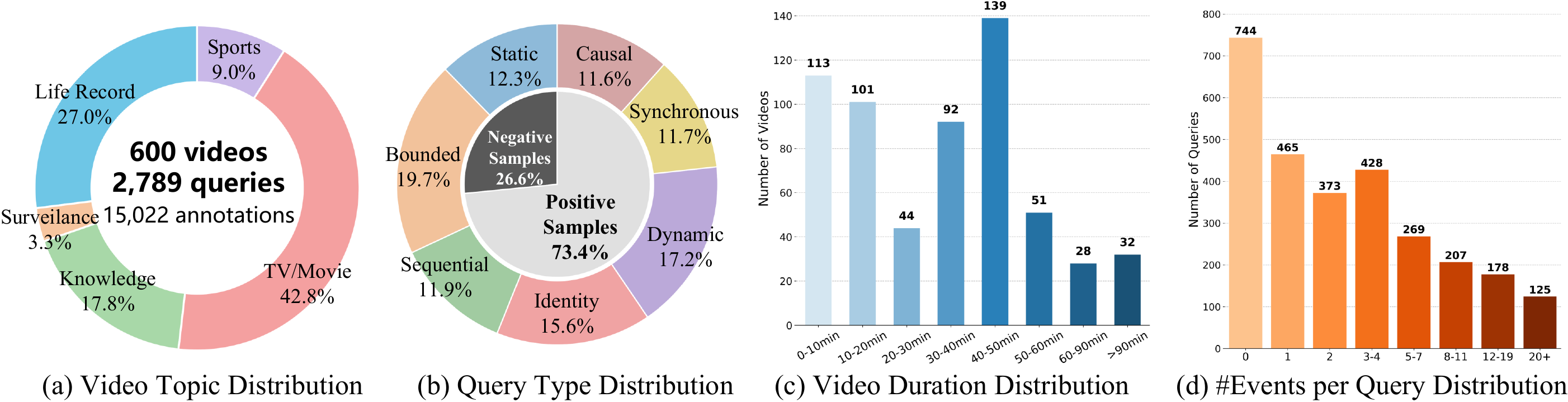}
  \caption{\textbf{Dataset statistics of CoMET-Bench.} \textbf{(a)}~Video distribution across five real-world domains. \textbf{(b)}~Distribution of query type occurrence: the inner ring splits positive and negative queries; the outer ring breaks positives into four temporal conditions (Causal, Sequential, Synchronous, Bounded) and three spatial conditions (Static, Dynamic, Identity). \textbf{(c)}~Video-duration distribution, ranging from under 10 minutes to over 90 minutes. \textbf{(d)}~Distribution of the number of grounded events per query, covering zero-instance negatives to dense multi-event cases.}
  \label{fig:data_statistics}
  \vspace{-6pt}

\end{figure*}

Fig.~\ref{fig:data_statistics} summarizes CoMET-Bench's diversity along four axes: video topic, query type, video duration, and event density. \textbf{(1)~Topic.} Videos span five real-world domains (Fig.~\ref{fig:data_statistics} (a)), led by TV/Movie (42.8\%) and Life Record (27.0\%), with Knowledge, Sports, and Surveillance covering the remaining 30.1\%. \textbf{(2)~Query type.} Among the 2{,}789 queries, 73.4\% are positive, and 26.6\% are negative (Fig.~\ref{fig:data_statistics} (b)), supporting joint evaluation of retrieval capability and hallucination resistance. Across the seven positive attributes, four temporal conditions (Causal 11.6\%, Sequential 11.9\%, Synchronous 11.7\%, Bounded 19.7\%) and three spatial conditions (Static 12.3\%, Dynamic 17.2\%, Identity 15.6\%) jointly span a broad compositional space. \textbf{(3)~Duration.} The 600 videos span from $<$10 to $>$90 minutes, with 111 videos over 50 minutes, including 32 over 90 minutes (Fig.~\ref{fig:data_statistics} (c)). The average duration of 33.8 minutes places CoMET-Bench firmly in the long-form regime. \textbf{(4)~Event density.} The number of target events per query ranges from 0 to 20+ (Fig.~\ref{fig:data_statistics} (d)), with a long tail of dense cases (510 queries with $\geq 8$ events) that prevents models from succeeding on sparse instances alone.

\subsection{Evaluation Metrics}
\label{evaluation metrics}
We evaluate CoMET-Bench along three aspects, each targeting a distinct failure mode of conditional multi-event temporal grounding. \textbf{(1) Multi-event counting} (\emph{MAE}, \emph{OBO}, \emph{Pearson}) measures whether the predicted instance count matches the ground truth. \textbf{(2) Temporal event grounding} (\emph{mIoU}, \emph{Recall@0.5}, \emph{F1@0.5}) measures whether predicted intervals align with ground-truth intervals. \textbf{(3) Negative query recognition} reports the proposed \textbf{Rejection-F1}, the harmonic mean of rejection recall on negatives and non-empty coverage on positives, which prevents trivial gaming by lazy ``always-empty'' models, alongside \textbf{FPR} as a direct readout of hallucination severity. Detailed formulations and motivations are provided in Appendix~\ref{sec:app_metric_formulations}.

\section{Methodology}
\label{method}

\begin{figure}[t]
  \centering
  \includegraphics[width=1\linewidth]{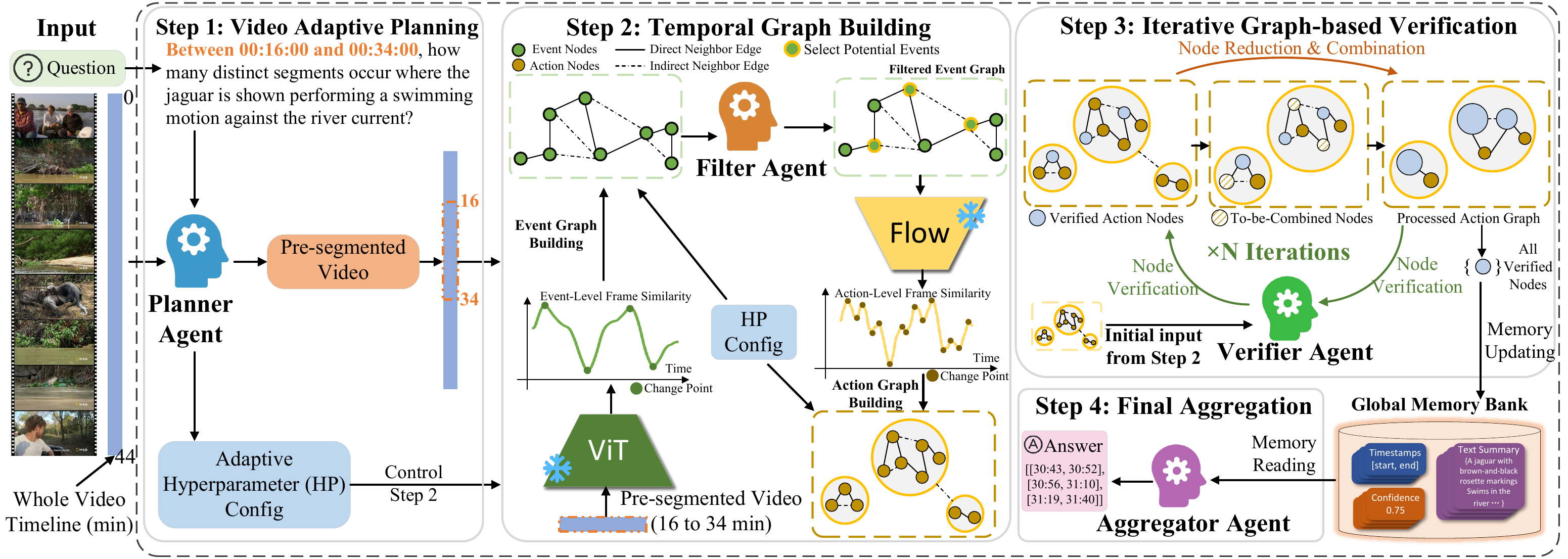}
  \caption{Overview of the proposed \textbf{CoMET-Agent} framework. Given a long-form video and a natural-language query, a single MLLM orchestrates four specialized agents (\emph{Planner}, \emph{Filter}, \emph{Verifier}, and \emph{Aggregator}) across four sequential steps to address conditional multi-event temporal grounding. Snowflake icons denote frozen (training-free) components.}
  \label{fig:method_pipeline}
  \vspace{-6pt}
\end{figure}

We propose \textbf{CoMET-Agent} (details are in Appendix \ref{appendix_method}), a training-free agentic framework that reformulates conditional multi-event temporal grounding as a \emph{search-and-aggregate} problem over a hierarchical video temporal graph with a persistent memory bank. Fig.~\ref{fig:method_pipeline} illustrates our four-step end-to-end pipeline orchestrated by a single MLLM. In \textbf{Step~1 (Video Adaptive Planning)}, the \emph{Planner Agent} parses the natural-language query to determine an adaptive hyperparameter (HP) configuration for subsequent graph construction. If the query specifies an explicit bounded temporal window, the agent also trims the video to a pre-segmented clip and passes it into Step 2 (Details in the Appendix \ref{sec:app_preseg}).
In \textbf{Step~2 (Temporal Graph Building)}, we construct a hierarchical video temporal graph. First, ViT features~\cite{simeoni2025dinov3} combined with a change-point detection algorithm~\cite{killick2012change_point_detection} on inter-frame similarity segment the video into coarse \emph{Event Nodes}. Next, a \emph{Filter Agent} screens out irrelevant events. The retained potential events are then expanded into fine-grained \emph{Action Nodes} via optical-flow segmentation~\cite{teed2020raft}.
In \textbf{Step~3 (Iterative Graph-based Verification)}, the \emph{Verifier Agent} iteratively traverses the action graph: in each iteration, it expands candidate nodes with their neighbors to enlarge the perception context, verifies whether the expanded context satisfies the query conditions, and commits every verified node to the \emph{Global Memory Bank} (Sec.~\ref{global memory bank}). In \textbf{Step~4 (Final Aggregation)}, the \emph{Aggregator Agent} consults the populated memory bank to produce the final grounded intervals and exact count.

\subsection{Hierarchical Temporal Graph Building}
\label{sec:graph_building}

To avoid the computational bottleneck of exhaustive frame-by-frame evaluation over long-form videos, we construct a hierarchical temporal graph that organizes the video into a searchable two-level structure: a coarse \emph{Event Graph} for rapid semantic localization, and a fine-grained \emph{Action Graph} for precise boundary delineation. Details are explained in Appendix \ref{sec:app_graph_edges}

\noindent \textbf{Coarse Event Graph Building.} We first construct the Event Graph $\mathcal{G}_E = (\mathcal{V}_E, \mathcal{E}_E)$, which serves as the coarse-grained search index over the video. To define the node set $\mathcal{V}_E$, we extract visual features using a pre-trained ViT model~\cite{simeoni2025dinov3} and apply a change-point detection algorithm~\cite{killick2012change_point_detection} on the inter-frame cosine similarities. This process segments the continuous video into $N$ discrete coarse-grained event nodes, where each $v_i \in \mathcal{V}_E$ represents a temporal segment $[t_i^s, t_i^e]$. The edge set $\mathcal{E}_E$ encodes the temporal relations that guide the search: edge weights are computed using a video embedding model~\cite{bai2025qwen3vl}, and we establish two types of edges to expose both local and long-range \emph{search paths}---\emph{direct neighbor edges} connecting strictly adjacent segments, and \emph{indirect neighbor edges} connecting non-adjacent segments within a bounded temporal distance. Once $\mathcal{G}_E$ is built, we deploy a \emph{Filter Agent} that prunes the search space: it evaluates the sampled frames of each event node against the query and discards clearly irrelevant background segments, yielding a focused set of potential events to be further searched at finer granularity.

\noindent \textbf{Fine-grained Action Graph Building.} While event nodes efficiently capture high-level semantic chunks, they are often too coarse to determine the exact action boundaries required for precise temporal grounding. Therefore, we expand the retained potential events into a fine-grained Action Graph, $\mathcal{G}_A = (\mathcal{V}_A, \mathcal{E}_A)$. Specifically, within each retained event node $v_i\in \mathcal{A}_E$, we extract frame-level dense optical flow using RAFT~\cite{teed2020raft} and compute the cosine similarities between adjacent flow representations. By applying the same change-point detection algorithm used for the Event Graph to these motion-based similarities, we segment $v_i$ into smaller, motion-homogeneous \emph{Action Nodes} $\mathcal{V}_A$. To maintain structural consistency, the edge set $\mathcal{E}_A$ inherits the direct and indirect connectivity logic similar in Event Graph. This fine-grained, interconnected topology serves as the foundational structure for the subsequent iterative verification process.

\subsection{Iterative Graph-based Verification}
\label{iterative graph-based verification}
While the fine-grained Action Graph $\mathcal{G}_A$ provides precise temporal boundaries, individual action nodes often lack the broader contextual information necessary for complex temporal reasoning (\emph{e.g.}, determining if an action occurred ``after a referee's red card''). To address this, we introduce an iterative, graph-based verification algorithm that dynamically expands the perception context before evaluating each candidate node. Detailed formulation could be found in Alg.~\ref{alg:verification} and \ref{sec:app_algorithm}.

\noindent \textbf{Node Reduction and Combination.} Instead of processing every node in isolation, we iteratively traverse the candidate nodes in $\mathcal{V}_A$. For a given candidate action node $v_i \in \mathcal{V}_A$, we expand its context by combining it with its topological neighborhood (comprising both direct and indirect neighbors) connected by edges whose values exceed a predefined threshold $\tau$, resulting in $\mathcal{C}(v_i)$. This aggregated node set is then forwarded to the \emph{Verifier Agent} for evaluation in the current iteration.

\noindent \textbf{Condition Verification and Memory Updating.} At each search step, the \emph{Verifier Agent} takes the sampled frames from the perception context $\mathcal{C}(v_i)$ and the query as input, and decides whether the candidate action node $v_i$ satisfies both the spatial semantics and the overarching temporal conditions specified in the query. A node passing this check is recorded as a verified hit and committed to the \emph{Global Memory Bank} (Sec.~\ref{global memory bank}); otherwise, the search proceeds to the next candidate without writing to memory. This selective writing ensures that the search outcome accumulated in memory contains only condition-compliant evidence, ready for downstream aggregation.

\subsection{Global Memory Bank}
\label{global memory bank}
To support global perception across the entire long-form video, we maintain a persistent \emph{Global Memory Bank} $\mathcal{M}$. Each verified node is committed as a tuple $m = \langle \mathcal{S}, (t^\mathrm{start}, t^\mathrm{end}), c \rangle$, where $\mathcal{S}$ is a textual summary, $(t^\mathrm{start}, t^\mathrm{end})$ are inherited timestamps, and $c \in [0,1]$ is the verifier's confidence. An online deduplication mechanism prevents redundant entries from neighboring nodes firing on the same event. In Step~4, the \emph{Aggregator Agent} consolidates $\mathcal{M}$ into the final answer and resolves identity consistency across temporally disjoint entries by cross-referencing their summaries~$\mathcal{S}$. Detailed mechanisms and formulations are in Appendix~\ref{sec:app_global_memory_bank}.

\section{Experiments}
\label{experiments}

\noindent\textbf{Experimental Setup.}
We benchmark CoMET-Bench across three representative families of video understanding methods. \emph{General-purpose Video MLLMs} include proprietary frontier models~\cite{gpt5, google_gemini3} and open-source models at 7--38B scales~\cite{bai2025qwen3vl, wang2025internvl3_5, llava-video, llavaov1.5, chen2025eagle2.5}, probing whether single-pass inference and model scale alone suffice. \emph{Agent-based methods}~\cite{yang2024doraemongpt, liu2025videomind, shen2025vgent, ye2025T*, yin2025videoarm} test whether existing decompose-and-retrieve pipelines transfer to compositional multi-event queries. \emph{Temporal-grounding-specialized models}~\cite{guo2024trace, zeng2025distime, zhang2025timelens, huang2024lita} assess whether grounding-specific training generalizes to our setting. All models use uniform sampling at 128 frames per video. CoMET-Agent is \textbf{training-free} and instantiated with two backbones (Qwen3-VL-8B~\cite{bai2025qwen3vl} and GPT5~\cite{gpt5}) to demonstrate portability across the open-source and proprietary regimes, paired with frozen Qwen3-VL-Embedding-8B~\cite{bai2025qwen3vl}, DINOv2-L/14~\cite{verified_dinov2}, and RAFT~\cite{teed2020raft} to build a temporal graph. More Details are in Appendix~\ref{implementation details}.


\begin{table}[t]
  \centering
  \scriptsize
  \setlength{\tabcolsep}{3pt}
  \renewcommand{\arraystretch}{0.7}
  \caption{Main results on CoMET-Bench across three evaluation aspects.
    \textbf{Counting}: MAE ($\downarrow$, mean absolute error), OBO ($\uparrow$, off-by-one accuracy), Pearson ($\uparrow$, Pearson correlation).
    \textbf{Grounding}: mIoU ($\uparrow$), R@0.5 ($\uparrow$, Recall at tIoU $0.5$), F1@0.5 ($\uparrow$).
    \textbf{Negative}: Rej.-F1 ($\uparrow$, harmonic mean of negative-rejection rate and positive-coverage rate), FPR ($\downarrow$, fraction of negative queries with non-empty prediction).
    \textbf{Bold} marks the best score per metric; \underline{underline} marks the second best.}
  \label{tab:main_results}
  \resizebox{\linewidth}{!}{%
  \begin{tabular}{l c ccc ccc cc}
    \toprule
    \multirow{2}{*}{\textbf{Method}} & \multirow{2}{*}{\textbf{\#Size}}
    & \multicolumn{3}{c}{\textbf{Counting}}
    & \multicolumn{3}{c}{\textbf{Grounding (Pos.\ only)}}
    & \multicolumn{2}{c}{\textbf{Negative}} \\
    \cmidrule(lr){3-5} \cmidrule(lr){6-8} \cmidrule(lr){9-10}
    & & MAE$\downarrow$ & OBO$\uparrow$ & Pear.$\uparrow$
    & mIoU$\uparrow$ & R@0.5$\uparrow$ & F1@0.5$\uparrow$
    & Rej.-F1$\uparrow$ & FPR$\downarrow$ \\
    \midrule
    \multicolumn{10}{l}{\textit{\textcolor{gray!90}{Video MLLMs}}} \\
    GPT-5~\cite{gpt5}                        & --   & 3.2 & 56.8 & 9.0 & 9.7  & 9.5  & 10.1 & 60.6 & \underline{5.3} \\
    Gemini~2.5~Pro~\cite{comanici2025gemini2.5}     & --   & 3.3 & 53.6 & 4.8 & 10.8 & 9.9  & 10.1 & 60.8 & 37.3 \\
    Gemini~3~Flash~\cite{google_gemini3}     & --   & 3.4 & 54.2 & 6.6 & 16.5 & 15.8 & 14.6 & 61.5 & 51.7 \\
    LLaVA-Video~\cite{llava-video}           & 7B   & 3.5 & 55.3 & 2.4 & 0.7  & 0.3  & 0.4  & 48.9 & 58.5 \\
    LLaVA-OV1.5~\cite{llavaov1.5}            & 7B   & 3.7 & 47.7 & 3.9 & 0.3  & 0.2  & 0.2  & 9.3  & \textbf{2.6} \\
    LongVILA-R1~\cite{chen2025longvila-r1}& 7B   & 5.7 & 38.2 & 11.1 & 10.8 & 7.6 & 4.6 & 5.6 & 97.1 \\
    MiMO-VL~\cite{coreteam2025mimovl}      & 7B   & 3.8 & 48.5 & -1.3 & 0.3 & 0.1 & 0.1 & 8.8  & 3.8 \\
    InternVL3.5~\cite{wang2025internvl3_5} & 8B  & 3.6 & 52.3 & 3.2 & 1.9 & 1.7 & 1.1 & 64.7 & 16.8 \\
    Eagle2.5~\cite{chen2025eagle2.5}         & 8B   & 3.9 & 48.8 & 1.0 & 2.1  & 0.9  & 0.7  & 39.9 & 14.6 \\
    Qwen3-VL~\cite{bai2025qwen3vl}           & 8.3B & 4.8 & 51.7 & 9.8 & 5.1  & 3.9  & 3.4  & 52.1 & \underline{5.3} \\
    Qwen3-VL~\cite{bai2025qwen3vl}     & 30B  & 3.6 & 51.5 & 3.4 & 3.3 & 3.0 & 3.1 & 47.2 & 5.9 \\
    InternVL3.5~\cite{wang2025internvl3_5} & 38B  & 4.0 & 48.3 & 13.4 & 2.6 & 1.8 & 1.3 & 40.3 & 9.1 \\
    \midrule
    \multicolumn{10}{l}{\textit{\textcolor{gray!90}{Agent-based Methods}}} \\
    VideoMind~\cite{liu2025videomind}     & 7B   & 4.4 & 14.8 & -0.5 & 7.6 & 4.8 & 2.3 & 1.2 & 99.4 \\
    Vgent-adapted~\cite{shen2025vgent} (Qwen2.5-VL) & 7B & 6.23 & 51.9 & 12.4 & 2.9 & 2.6 & 2.0 & 48.5 & 5.4 \\
    T$^*$-adapted~\cite{ye2025T*} (GPT-4o) & -- & 3.80 & 44.2 & 16.3 & 1.2 & 1.0 & 1.4 & 34.3 & 0.0 \\
    TimeSearch-R~\cite{verified_timesearch_r} (Qwen2.5-VL) & 7B & 3.30 & 48.4 & -0.3 & 8.4 & 5.5 & 5.0 & 26.1 & 83.7 \\
    VideoARM~\cite{yin2025videoarm} (GPT-4.1\&o3) & -- & 3.6 & 52.1 & \textbf{14.5} & 13.5 & 12.8 & 13.0 & 61.0 & 9.5 \\
    \midrule
    \multicolumn{10}{l}{\textit{\textcolor{gray!90}{Temporal Grounding-Specialized}}} \\
    TRACE~\cite{guo2024trace}                & 7B   & 3.2 & 60.9 & 1.0 & 4.0  & 3.3  & 3.8  & 2.5  & 98.8 \\
    DisTime~\cite{zeng2025distime}    & 7B   & 3.3 & 60.0 & -0.6 & 2.7 & 0.5 & 0.5 & 12.1 & 93.5 \\
    TimeLens~\cite{zhang2025timelens}  & 7B   & 3.3 & 60.7 & 0.0 & 6.2 & 5.0 & 5.4 & 0.0 & 100.0 \\
    LITA~\cite{huang2024lita}                & 13B  & 3.7 & 58.4 & 0.2 & 1.8  & 0.4  & 0.4  & 1.7  & 99.1 \\
    \midrule
    \rowhl
    \textbf{Ours (Qwen3-VL)}                 & 8.6B & 3.9 & 54.5 & 11.0 & 9.2  & 8.5  & 8.8  & 57.4 & 8.0 \\
    \rowhl
    \textbf{Ours (GPT-5)}                    & --   & \underline{2.8} & \underline{61.4} & \underline{13.7} & \underline{17.5} & \underline{17.6} & \underline{16.2} & \underline{66.5} & 6.4 \\
    \rowhl
    \textbf{Ours (Gemini 3 Flash)}           & --   & \textbf{2.7} & \textbf{62.0} & 12.0 & \textbf{19.3} & \textbf{19.5} & \textbf{19.0} & \textbf{68.1} & 5.8 \\
    \bottomrule
  \end{tabular}%
  }
\end{table}

\noindent\textbf{Completed agent-baseline evaluation.}
Vgent-adapted and T$^*$-adapted retain their original search procedures but emit interval sets; Vgent's terminal generation budget is increased from 10 to 2,048 tokens. Their final F1@0.5 values are 2.0\% and 1.4\%, respectively; TimeSearch-R reaches 5.0\%. All three runs cover 2,789 queries and replace the early partial evaluations. Low FPR alone does not imply successful grounding: T$^*$-adapted has FPR 0.0\%, but Rejection-F1 34.3\% and F1@0.5 1.4\%.

\noindent\textbf{Main Results.}
Tab.~\ref{tab:main_results} reports performance across counting, grounding, and negative-query metrics, from which we highlight six findings:
\textbf{(1)} CoMET-Agent lifts GPT-5 from F1@0.5 10.1\% to 16.2\% and Gemini~3~Flash from 14.6\% to 19.0\%, surpassing all single-pass baselines purely through framework design.
\textbf{(2)} Scaling Qwen3-VL from 8.3B to 30B does not help and even hurts grounding (F1@0.5 3.4\% $\to$ 3.1\%), suggesting that raw parameter count is not the bottleneck.
\textbf{(3)} Proprietary frontier models substantially outperform open-source ones on grounding (F1@0.5 10.1\%--14.6\% vs $\leq$4.6\% across all 7--38B open-source MLLMs).
\textbf{(4)} Several agent-based methods (VideoMind, Vgent, T$^*$) either fail or cannot follow our task instruction, exposing a brittleness of decompose-and-retrieve pipelines on multi-event queries (Appendix~\ref{agent method benchmarking}).
\textbf{(5)} Temporal Grounding-specialized models collapse on our task: F1@0.5 below 5.4\% and FPR above 93\%, indicating that grounding-tuned training does not transfer to a compositional multi-event setting and severely amplifies hallucination.
\textbf{(6)} The proposed \textbf{Rejection-F1} successfully exposes ``always-empty'' models like LLaVA-OV1.5 and MiMO-VL attain the lowest FPR yet collapses on Rej.-F1, confirming that FPR or rejection rate alone is gameable while Rej.-F1 is not.
Note that more in-depth analyses are in Appendix~\ref{additional analysis benchmark}.

\begin{figure}[t]
  \centering
  \begin{subfigure}[t]{0.45\linewidth}
    \centering
    \includegraphics[width=\linewidth]{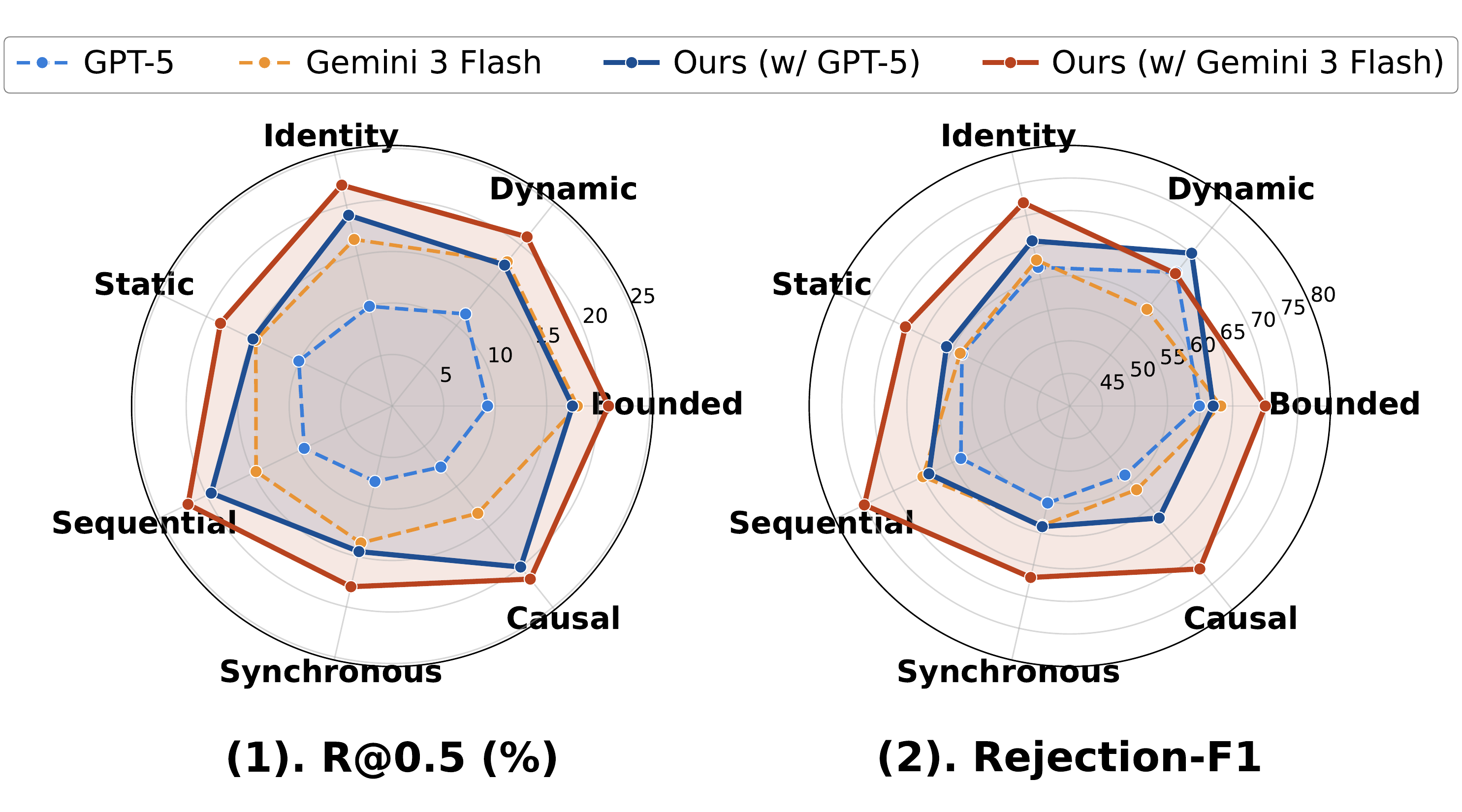}
    \caption{Per-query type R@0.5 and Rej.-F1.}
    \label{fig:radar_attribute}
  \end{subfigure}
  \hfill
  \begin{subfigure}[t]{0.54\linewidth}
    \centering
    \includegraphics[width=\linewidth]{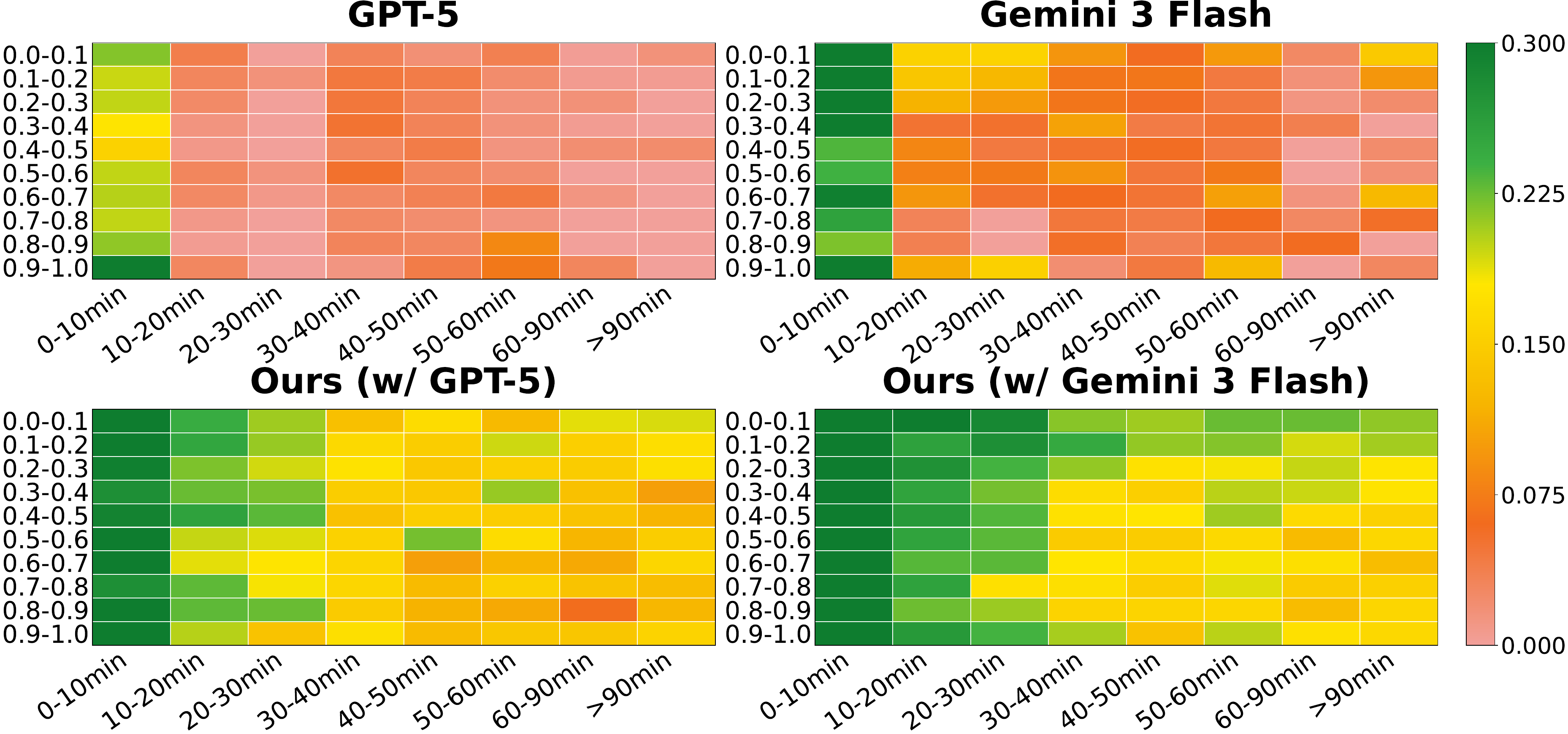}
    \caption{mIoU \emph{w.r.t.} video duration and event position.}
    \label{fig:position_heatmap}
  \end{subfigure}
  \caption{\textbf{Where CoMET-Agent's gains come from.} \textbf{(a)} Attribute-level R@0.5 and Rej.-F1. \textbf{(b)} mIoU heatmap with respect to (\emph{w.r.t.}) video duration and ground-truth moment position.}
  \label{fig:analysis}
  \vspace{-5pt}
\end{figure}

\noindent\textbf{Why does CoMET-Agent help?}
\textbf{(1) Compositional temporal conditions.}
As shown in Fig.~\ref{fig:analysis}, best-performing MLLMs collapse on cross-moment attributes (\emph{Sequential}, \emph{Synchronous}, \emph{Causal}; R@0.5 below 8\% for GPT-5), and CoMET-Agent's largest gains concentrate exactly there (Sequential R@0.5: 7\%~$\to$~22\% under the same GPT-5 backbone), showing the iterative graph verification and memory bank target compositional temporal reasoning rather than perception.
\textbf{(2) Long-form videos.} (Fig. \ref{fig:position_heatmap})
Both baselines collapse to near-zero mIoU once the video exceeds 10 minutes, with failures uniform across all moment positions, indicating evidence dilution rather than a middle-position bias. CoMET-Agent retains non-trivial mIoU beyond 90 minutes and across all positions.
\noindent\textbf{Reported inference latency.}
The completed comparison in Tab.~\ref{tab:latency_rebuttal} reports 27.5 s/query for CoMET-Agent (Qwen2.5-VL), versus 49.1 for TimeSearch-R and 87.0 for Vgent-adapted. With GPT-4o, CoMET-Agent takes 21.3 s/query and T$^*$-adapted 207.8 s/query, a reported ratio of $9.8\times$. These are wall-clock measurements under the reported hardware/API settings, not hardware-normalized complexity guarantees.

\textbf{(3) Efficiency.}
CoMET-Agent adds only $\sim$0.3B frozen parameters (DINOv2 and RAFT). Additional results, ablations, and efficiency analysis are in Appendix~\ref{method performance}, \ref{method ablations}~and~\ref{method efficiency}).

\begin{figure}
    \centering
    \includegraphics[width=1\linewidth]{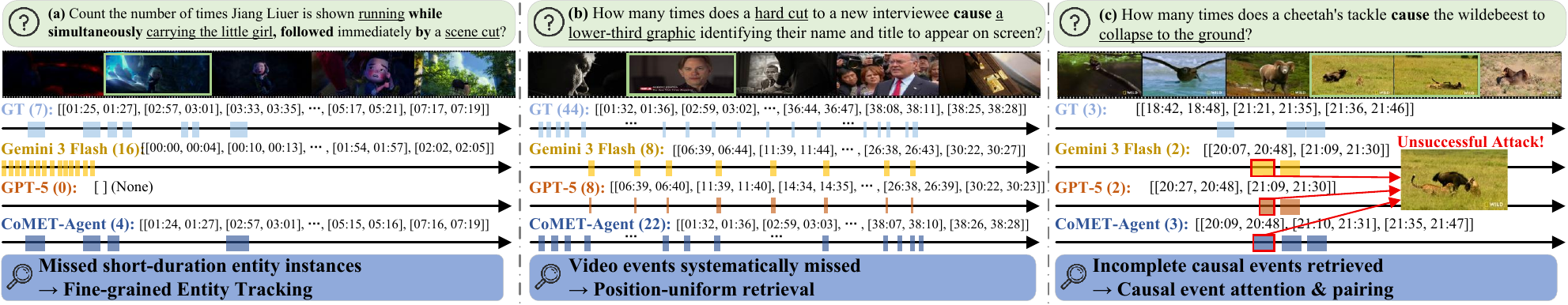}
\caption{\textbf{Failure cases analysis of CoMET-Agent, each motivating a future direction.}}
\vspace{-6pt}
\label{fig:failure_cases}
\end{figure}

\noindent\textbf{Unsolved Aspects of CoMET-Agent.}
Despite consistent gains, three regimes remain challenging (Fig.~\ref{fig:failure_cases}).
\textbf{(1) Fine-grained entity tracking} (Fig.~\ref{fig:failure_cases}(a)): when target events are short and feature visually similar co-occurring entities, the verifier's textual identity summary becomes ambiguous and CoMET-Agent under-counts, motivating dedicated object trackers~\cite{ravi2024sam2, yang2024samurai} integrated into our search-and-aggregate pipeline.
\textbf{(2) Position-uniform retrieval} (Fig.~\ref{fig:failure_cases}(b)): on densely distributed events across long videos, CoMET-Agent recalls only 22 of 44 instances and the remaining errors concentrate at mid- and late-video positions, calling for coverage-aware traversal that explicitly enforces uniform exploration along the timeline.
\textbf{(3) Causal event pairing} (Fig.~\ref{fig:failure_cases}(c)): all baselines including CoMET-Agent ground visually similar but non-causal instances (e.g., \emph{unsuccessful attack} that never lead to a collapse), suggesting an explicit cause-effect attention mechanism that pairs the cause and the consequence (\emph{tackle} $\to$ \emph{collapse}) across the action graph as a promising direction. We further discuss limitation and future works in Appendix~\ref{sec:limitations_future}.
\vspace{-2pt}

\section{Conclusion}
We introduced \textbf{CoMET-Bench}, the first benchmark for Conditional Multi-Event Temporal Grounding in long-form video, together with a unified evaluation protocol and the proposed Rejection-F1 metric. Benchmarking reveals that no existing method handles compositional multi-event queries well. We further proposed \textbf{CoMET-Agent}, a training-free agentic framework that substantially improves over single-pass inference, and identify entity tracking, position-uniform retrieval, and causal event pairing as open directions for future work.

\clearpage
\begin{ack}
This work was supported in part by a gift from Qualcomm. This work also used the Delta system at the National Center for Supercomputing Applications, which is supported by the National Science Foundation under Award No. OAC-2005572, through allocations CIS260306 and CIS250828 from the Advanced Cyberinfrastructure Coordination Ecosystem: Services \& Support (ACCESS) program. ACCESS is supported by the U.S. National Science Foundation under Grants No. 2138259, 2138286, 2138307, 2137603, and 2138296.

The authors thank Francisco Arisso, Ayaan Khan, David Orjuela, Harshitha Sathees Kumar, and Jeremy Triana for their contributions to the annotation of the data.
\end{ack}

{\small
  \bibliographystyle{plain}
  \bibliography{references}
}

\newpage
\appendix

\section*{Appendix Overview}
This appendix is organized as follows:
\begin{itemize}[leftmargin=1.5em]
  \item \textbf{A.} \hyperref[additional related works]{\textbf{Related Works of Video Understanding Methods}}
  \item \textbf{B.} \hyperref[sec:app_data_annotation]{\textbf{Additional Details of Benchmark Construction}}
  \item \textbf{C.} \hyperref[appendix_method]{\textbf{Additional Details of Methodology}}
  \item \textbf{D.} \hyperref[implementation details]{\textbf{Implementation Details}}
  \item \textbf{E.} \hyperref[additional experiments]{\textbf{Additional Experiments}}
  \item \textbf{F.} \hyperref[sec:limitations_future]{\textbf{Challenges and Future Work}}
\end{itemize}

\section{Related Works of Video Understanding Methods}
\label{additional related works}

Existing video understanding approaches fall into two categories. \textbf{(i)~Video MLLMs}---including general-purpose models~\cite{gpt5, google_gemini3, bai2025qwen3vl, wang2025internvl3_5, llavaov1.5, coreteam2025mimovl}, VTG-specialized variants such as TimeChat~\cite{ren2024timechat} (time-aware encoder), VTimeLLM~\cite{huang2024vtimellm} (normalized timestamp tokens), LITA~\cite{huang2024lita} (SlowFast tokens for reasoning temporal localization), TRACE~\cite{guo2024trace} (causal event modeling), and Number-It~\cite{wu2025number_it} (numerical frame identifiers), and memory- or context-extension methods~\cite{song2024moviechat, he2024ma-lmm, zhang2025flashvstream, shen2024longvu, chen2024longvila}---all process the video in a single forward pass and predict one best-matching moment per query. \textbf{(ii)~Agent-based methods} decompose the task into structured sub-routines: VideoAgent~\cite{wang2024videoagent} iteratively retrieves key frames via CLIP-guided tool use; VideoTree~\cite{wang2025videotree} constructs query-adaptive hierarchical trees through visual clustering; DoraemonGPT~\cite{yang2024doraemongpt} converts videos into SQL-queryable symbolic memory; VideoMind~\cite{liu2025videomind} trains Planner/Grounder/Verifier/Answerer roles to complete each sub-task for video understanding; Vgent~\cite{shen2025vgent} represents videos as entity-centric graphs for retrieval-augmented generation; and T*~\cite{ye2025T*} explicitly frames long-video understanding as iterative temporal search with adaptive zoom-in. Across both categories, however, grounding is treated as \emph{retrieval}---returning a single best moment or a sparse set of relevant frames---and none provides a quantitative aggregation mechanism for exhaustive multi-instance counting under compositional temporal conditions, leaving the gap that CoMET-Agent addresses.

\section{Additional Details of Benchmark Construction}
\label{sec:app_data_annotation}

This appendix provides additional details on the construction of CoMET-Bench, complementing the high-level description in Sec.~\ref{data construction}. We first present the end-to-end annotation pipeline (Sec.~\ref{sec:app_pipeline_overview}), followed by the prompts used for MLLM-assisted query generation (Sec.~\ref{sec:app_query_prompt}) and annotation generation (Sec.~\ref{sec:app_annotation_prompt}). Finally, we describe the human verification protocol and the quality-control statistics (Sec.~\ref{sec:app_human_verification}).

\subsection{Copyright and License Clarification}
\label{sec:app_copyright}

The 600 videos in CoMET-Bench are sourced from three publicly released long-video benchmarks and a small set ($\leq 20$) of YouTube videos collected by us. \textbf{Video-MME}~\cite{data-videomme} is released for academic research only with copyright held by the original video owners; \textbf{MLVU}~\cite{data-zhou2025mlvu} is released under CC-BY-NC-SA-4.0 for non-commercial research; and \textbf{CG-Bench}~\cite{chen2024cg-bench-data} is distributed as a gated dataset for approved researchers only, and we accessed it through the official approval process. We use all videos strictly for non-commercial academic research, in compliance with their original terms. For the YouTube subset, we do not redistribute raw video files but release only video identifiers so that researchers can fetch them directly from the original sources, subject to YouTube's Terms of Service and the uploaders' copyright.

\begin{figure}[h]
  \centering
  \includegraphics[width=0.8\linewidth]{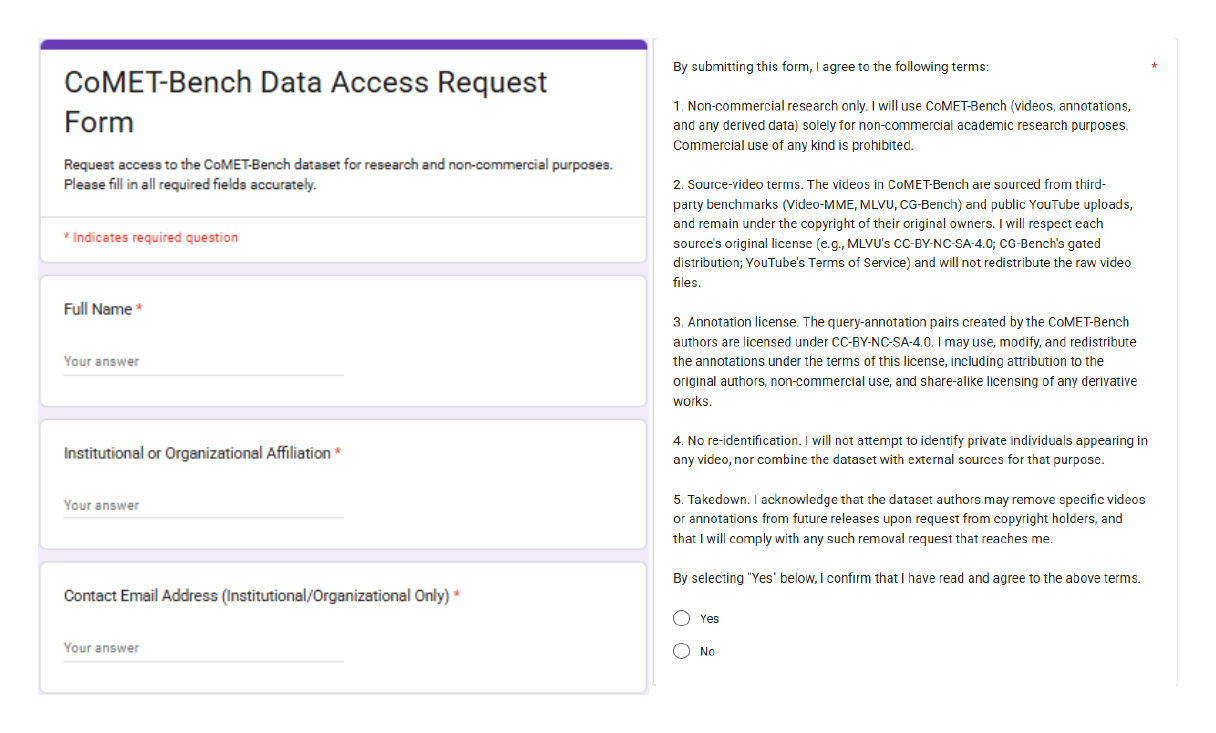}
  \caption{\textbf{CoMET-Bench data access request form.} Researchers must provide institutional credentials and explicitly agree to non-commercial academic use before being granted access to the dataset.}
  \label{fig:access_request_form}
\end{figure}

The query-annotation pairs created in this work, including natural-language queries, attribute tags, grounded intervals, and verification metadata, are released under CC-BY-NC-SA-4.0 to remain compatible with the most restrictive upstream source. To preserve the access controls inherited from CG-Bench and to ensure non-commercial academic use, CoMET-Bench is itself distributed as a \textbf{gated dataset}: researchers request access through the form shown in Fig.~\ref{fig:access_request_form}, providing their full name, institutional affiliation, and an institutional email, and explicitly agreeing not to redistribute, resell, or commercially exploit any portion of the dataset or its annotations. We respect the rights of all original video creators and will promptly remove any entries upon request from copyright holders through the project page.

\subsection{Annotation Pipeline Overview}
\label{sec:app_pipeline_overview}
\begin{figure}[t]
  \centering
  \includegraphics[width=1\linewidth]{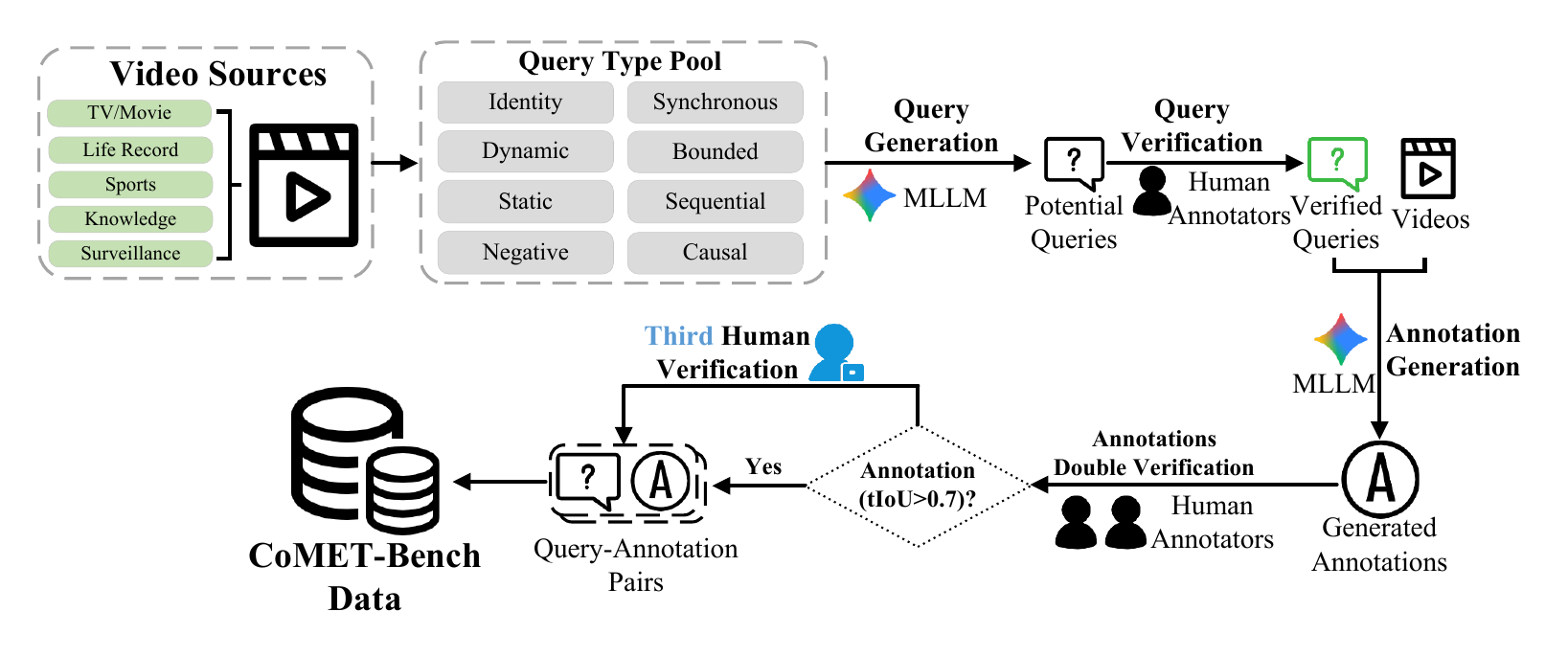}
  \caption{\textbf{Data construction pipeline of CoMET-Bench.}}
\label{fig:construction_pipeline}
\end{figure}

The detailed annotation pipeline is illustrated in Fig.~\ref{fig:construction_pipeline}. The pipeline consists of five stages. \textbf{(1)~Video collection}: 600 videos are curated from five real-world domains (TV/Movie, Life Record, Sports, Knowledge, Surveillance). \textbf{(2)~Query generation}: an MLLM generates candidate queries under attribute-driven prompting, where each query targets a specific combination drawn from the eight query attributes (four temporal: \emph{Causal}, \emph{Sequential}, \emph{Synchronous}, \emph{Bounded}; three spatial: \emph{Static}, \emph{Dynamic}, \emph{Identity}; and the orthogonal \emph{Negative} attribute). \textbf{(3)~Query verification}: human annotators verify each query for video relevance, naturalness, and real-world plausibility, rewriting or discarding those that fail. \textbf{(4)~Annotation generation}: the MLLM produces an initial set of grounded intervals for each verified query. \textbf{(5)~Two-pass annotation verification}: a pair of human annotators independently inspects each query--annotation pair against the source video, correcting boundaries, adding missed instances, and removing spurious ones; pairs reaching inter-annotator agreement above $\mathrm{IoU} > 0.7$ proceed to a final third-pass review by a senior annotator, while disagreements are adjudicated. Only pairs passing all three rounds of verification are included in the final benchmark.

\subsection{Query Generation}
\label{sec:app_query_prompt}
The query generation prompt instructs the MLLM (in practical we use Gemini 3 Pro-Preview \cite{google_gemini3}) to produce queries that are visually grounded, temporally precise, and compositional. To achieve compositionality, we sample attribute combinations from the eight-attribute pool before invoking the MLLM, and pass the selected combinations into the prompt as explicit constraints. Each combination contains 1--3 attributes, with combinations of 2--3 attributes oversampled (peak at $k=2.5$, Gaussian-weighted) to encourage compositional queries while still preserving simpler single-attribute cases. The MLLM is then instructed to generate exactly one query per combination, ensuring uniform coverage of the compositional query space.

For each video, we sample five attribute combinations using a biased random procedure: the combination size $k$ is drawn from $\{1, 2, \dots, 7\}$ with probability $p(k) \propto \exp(-\alpha (k - \mu)^2)$ where $\mu = 2.5$ and $\alpha = 0.9$, peaking at compositional queries of 2--3 attributes; the $k$ attributes are then sampled uniformly without replacement from the 8-attribute pool. This procedure ensures uniform coverage of the compositional space while concentrating effort on the queries with most diagnostic value.
 
\begin{promptbox}{Query Generation Prompt}
\small
\textbf{Task Overview:} You are an expert annotator building a benchmark for Conditional Multi-Event Temporal Grounding. Your goal is to generate 5 meaningful queries based on the video content provided.
 
\vspace{4pt}
\textbf{*** CRITICAL CONSTRAINT 1: VISUAL GROUNDING ***}\\
The queries MUST describe events that are VISUALLY OBSERVABLE in the video player pixel-by-pixel.
\begin{itemize}[leftmargin=*, itemsep=0pt, topsep=2pt]
  \item \textbf{BAD Query} (Abstract): \emph{``Count how many times the empire expands.''} (Too vague.)
  \item \textbf{GOOD Query} (Visual): \emph{``Count how many times the map border visually shifts or changes color.''} (Observable.)
\end{itemize}
 
\vspace{4pt}
\textbf{*** CRITICAL CONSTRAINT 2: STRICT TIME FORMAT (HH:MM:SS) ***}
\begin{itemize}[leftmargin=*, itemsep=0pt, topsep=2pt]
  \item You MUST use the 3-part HH:MM:SS format for ALL time references.
  \item NEVER use MM:SS (e.g., \texttt{12:30} is ambiguous).
  \item If the video is within 1 hour, write \texttt{00:MM:SS}.
  \item Examples: 30 seconds $\to$ \texttt{00:00:30}; 1 minute 5 seconds $\to$ \texttt{00:01:05}; 1 hour 15 minutes $\to$ \texttt{01:15:00}.
\end{itemize}
 
\vspace{4pt}
\textbf{Core Attribute Definitions:}
\begin{enumerate}[leftmargin=*, itemsep=0pt, topsep=2pt]
  \item \textbf{Identity}: Distinguish unique visual entities (e.g., \emph{``The man in the blue hat''}).
  \item \textbf{Static}: State-based attributes (e.g., \emph{``The map is entirely red''}).
  \item \textbf{Dynamic}: Motion analysis (e.g., \emph{``The hand shuffles the cards''}).
  \item \textbf{Causal}: A visual trigger leads to a visual reaction.
  \item \textbf{Sequential}: Visual events in order.
  \item \textbf{Synchronous}: Two things happening at the exact same time.
  \item \textbf{Bounded}: Constrains the counting to a specific timestamp range.
\end{enumerate}
 
\vspace{4pt}
\textbf{Query Requirements:}
\begin{itemize}[leftmargin=*, itemsep=0pt, topsep=2pt]
  \item If a combination has 2+ attributes, use at least two.
  \item Each query must specify exactly one target event type to count.
  \item STRICTLY AVOID spatial counting (e.g., \emph{``how many birds in the sky''}).
  \item At least 2 of the 5 queries must include a specific time window.
\end{itemize}
 
\vspace{4pt}
\textbf{=== FEW-SHOT EXAMPLES ===}
 
\textbf{Example 1} (Dynamic + Identity + Bounded)\\
\textit{Question:} Between 00:00:00 and 00:00:30, how many times does Jennifer Lawrence (Identity) perform a riffle shuffle (Dynamic) with the cards?\\
\textit{Target:} Count the specific ``shuffling'' motion visible on the FaceTime screen.
 
\vspace{2pt}
\textbf{Example 2} (Causal + Static)\\
\textit{Question:} Count the number of times a question text appears on screen (Causal) that immediately cuts to a static image of Jennifer looking surprised (Static).\\
\textit{Target:} Ground the cut between the text frame and the reaction frame.
 
\vspace{2pt}
\textbf{Example 3} (Sequential + Dynamic)\\
\textit{Question:} After the text ``Jack of Clubs'' appears, how many times is a card physically flipped by a hand?\\
\textit{Target:} Count each individual hand movement that changes the position of a card.
 
\vspace{2pt}
\textbf{Example 4} (Bounded)\\
\textit{Question:} From 00:45:00 to 00:46:30, how many times does the red arrow appear on the map?\\
\textit{Target:} Visual tracking of the red arrow graphic within the 90-second window.
 
\vspace{4pt}
\textbf{Attribute Combinations to use:}\\
\texttt{\{combinations\_block\}}
 
\vspace{4pt}
\textbf{=== OUTPUT FORMAT ===}\\
Return strictly valid JSON. No markdown. \texttt{"annotations"} must be empty.
 
\begin{verbatim}
{
  "queries": [
    {
      "query_text": "Question...",
      "annotations": []
    }
  ]
}
\end{verbatim}
\end{promptbox}

\subsection{Annotation Generation}
\label{sec:app_annotation_prompt}

For each verified query, we prompt a frontier MLLM to produce an initial set of grounded intervals as a starting point for human refinement. The prompt enforces three properties critical to our task: (i)~\emph{exhaustive enumeration}---every occurrence of the target pattern must be returned as a separate interval, ensuring multi-instance coverage rather than only the most salient match; (ii)~\emph{boundary strictness}---each interval starts when the target event first becomes visually identifiable and ends when it concludes or leaves the frame, providing tight grounding that downstream tIoU metrics can meaningfully evaluate; and (iii)~\emph{strict temporal format}---all timestamps follow the unambiguous \texttt{HH:MM:SS} convention. We additionally request a short \emph{descriptive evidence} note for each interval, which serves as an audit trail that human verifiers use during the subsequent refinement and adjudication stages. We emphasize that the MLLM output is treated as a strong candidate to be reviewed and corrected by human annotators, not as ground truth: human verifiers routinely correct misaligned boundaries, add missed instances, and remove spurious ones during the two-pass verification described in Sec.~\ref{data construction}.

\begin{promptbox}{Annotation Generation Prompt}
\small
\textbf{Role:} You are a Professional Video Annotation Specialist specializing in Temporal Grounding. Your goal is to produce high-fidelity datasets by mapping the given query to precise video intervals.

\vspace{4pt}
\textbf{Input:}
\begin{itemize}[leftmargin=*, itemsep=0pt, topsep=2pt]
  \item Given Video
  \item Query: \texttt{\{query\_text\}}
\end{itemize}

\vspace{4pt}
\textbf{Execution Protocol:}
\begin{enumerate}[leftmargin=*, itemsep=2pt, topsep=2pt]
  \item \textbf{Temporal Precision (HH:MM:SS):}
    \begin{itemize}[leftmargin=*, itemsep=0pt, topsep=2pt]
      \item You MUST use the 3-part \texttt{HH:MM:SS} format for ALL time references.
      \item NEVER use \texttt{MM:SS} (e.g., \texttt{12:30} is ambiguous).
      \item If the video is within 1 hour, write \texttt{00:MM:SS}.
      \item Examples: 30 seconds $\to$ \texttt{00:00:30}; 1 minute 5 seconds $\to$ \texttt{00:01:05}; 1 hour 15 minutes $\to$ \texttt{01:15:00}.
    \end{itemize}
  \item \textbf{Exhaustive Retrieval:} No occurrence of the target pattern should be omitted. If the pattern repeats 10 times, you must provide 10 distinct annotation blocks.
  \item \textbf{Boundary Strictness:} The \emph{start} is the moment the action/object becomes first identifiable. The \emph{end} is the moment the action concludes, or the object leaves the frame.
  \item \textbf{Descriptive Evidence:} In the \texttt{notes} field, describe the specific visual or auditory anchor that confirms the match.
\end{enumerate}

\vspace{4pt}
\textbf{OUTPUT FORMAT} (STRICT JSON ONLY, NO MARKDOWN):
\begin{verbatim}
{
  "annotations": [
    {
      "start_timestamp": "00:15:23",
      "end_timestamp": "00:15:29",
      "notes": "Short evidence note"
    }
  ]
}
\end{verbatim}
\end{promptbox}


\subsection{Human Verification}
\label{sec:app_human_verification}

\begin{figure}[ht]
  \centering
  \includegraphics[width=1\linewidth]{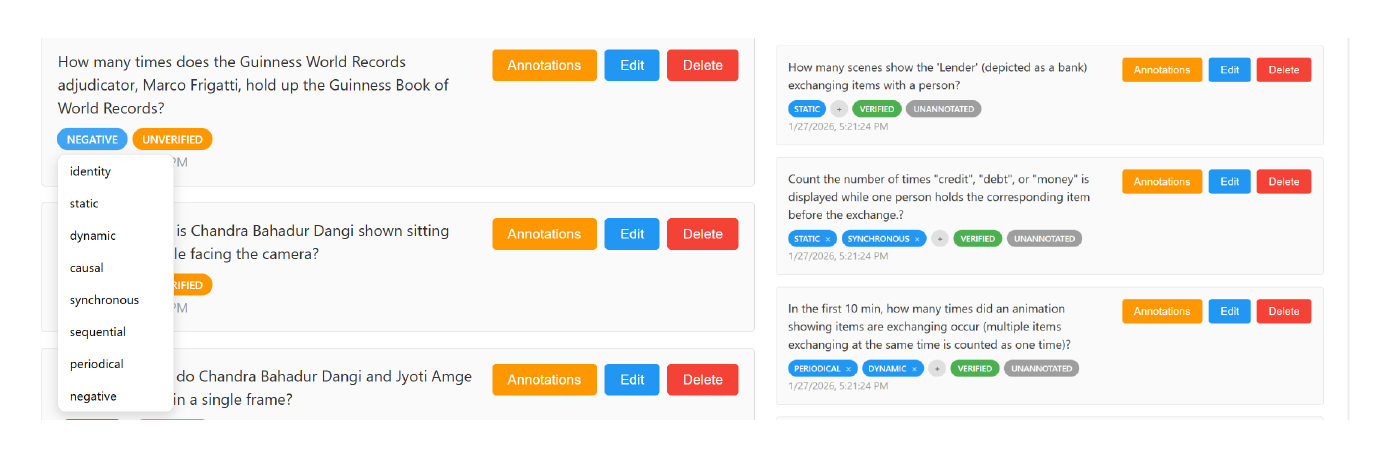}
  \caption{\textbf{Query verification interface of the CoMET-Bench annotation tool.} 
  For each candidate query produced by the MLLM, verifiers can assign one or more attribute tags from a dropdown (\emph{Identity}, \emph{Static}, \emph{Dynamic}, \emph{Causal}, \emph{Sequential}, \emph{Synchronous}, \emph{Bounded}, and \emph{Negative}), edit the query text, mark it as \texttt{Verified}, or discard it. Each query also displays its current verification and annotation status (\texttt{Verified}/\texttt{Unverified}, \texttt{Annotated}/\texttt{Unannotated}), allowing annotators to track progress at a glance.}
  \label{fig:tool_query_verification}
\end{figure}

\begin{figure}[ht]
  \centering
  \includegraphics[width=1\linewidth]{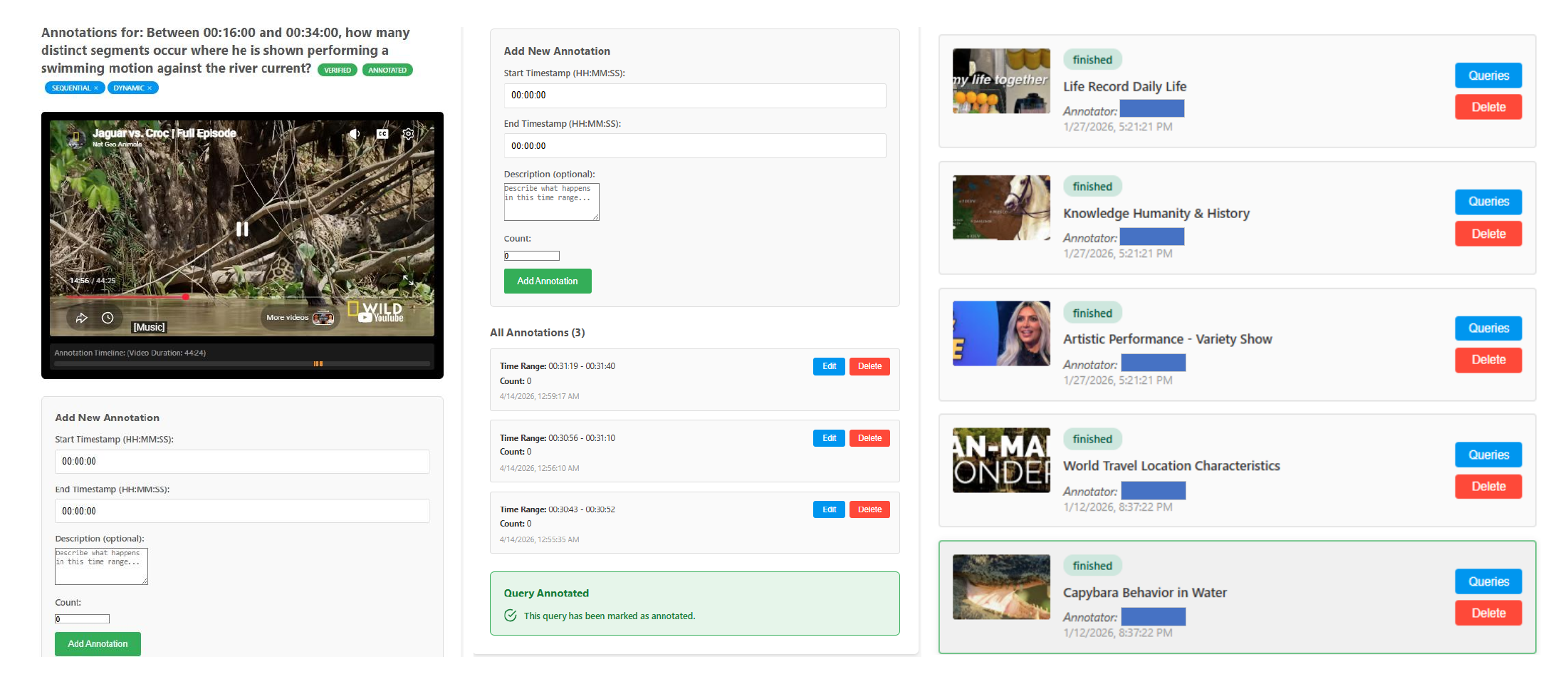}
  \caption{\textbf{Annotation verification interface of the CoMET-Bench annotation tool.} 
  Given a verified query and its source video (left), annotators inspect every MLLM-generated candidate interval shown in the central panel. Each interval can be edited (boundary correction in \texttt{HH:MM:SS} format), deleted (when spurious), or supplemented with new intervals to capture instances missed by the MLLM. A \texttt{Query Annotated} flag is set once the annotator confirms the interval set is exhaustive and correct. The right panel provides a video-level navigator across all assigned topics for tracking annotation progress.}
  \label{fig:tool_annotation_refinement}
\end{figure}

To ensure data quality, every CoMET-Bench query and its grounded annotations pass through a multi-stage verification protocol. We recruit trained annotators with backgrounds in computer vision and video annotation, and equip them with a custom web-based annotation interface (Fig.~\ref{fig:tool_query_verification} and \ref{fig:tool_annotation_refinement}) that integrates query refinement, attribute tagging, timestamp correction, and progress tracking into a single workflow. The full protocol consists of three stages, described below.

\textbf{Stage 1: Query Verification.}
For each video, the MLLM produces five candidate queries, and human verifiers examine them along three axes: \emph{understandability} (the query reads naturally and unambiguously), \emph{meaningfulness} (the query targets information someone would realistically want to extract), and \emph{difficulty} (the query is sufficiently challenging for current MLLMs). Verifiers are explicitly instructed to evaluate the query against three representative real-world application scenarios: (i)~\emph{video cutting} (finding repeated or highlighted patterns), (ii)~\emph{event counting} (finding complex, semantically meaningful events with clear visual features), and (iii)~\emph{record analysis} (locating recurring patterns of a specific player or interviewee, such as successful shots or mentions of a keyword). A query is considered well-formed if it reflects one or more of the eight task attributes (\emph{Identity}, \emph{Static}, \emph{Dynamic}, \emph{Causal}, \emph{Sequential}, \emph{Synchronous}, \emph{Bounded}, and \emph{Negative}). For each accepted query, verifiers assign the appropriate attribute tags (a query may carry multiple) and mark it as \texttt{Verified}. To preserve the temporal nature of the benchmark, verifiers are required to reject queries that reduce to \emph{spatial counting}, in which the model would simply count co-occurring objects within a single frame. Queries failing any of the above criteria are either rewritten or discarded before entering the annotation stage. Fig.~\ref{fig:tool_query_verification} shows the query-verification view of the annotation interface.

\textbf{Stage 2: Annotation Refinement.}
For each verified query, the MLLM produces a candidate set of grounded intervals, which is then refined by a human annotator. The annotator inspects the video against each candidate interval and performs four checks:
(i)~\emph{format check}---all timestamps follow the unambiguous \texttt{HH:MM:SS} convention;
(ii)~\emph{content check}---the visual content within each interval indeed satisfies the query;
(iii)~\emph{completeness check}---the annotator quickly scrubs through the entire video to identify qualifying instances missed by the MLLM;
(iv)~\emph{boundary correction}---intervals that only partially satisfy the query are tightened, extended, or split as appropriate.
Candidates that cannot be salvaged are deleted, and missed instances are added as new intervals. We deliberately leave the predicted count from the MLLM unverified at this stage, since the count is determined by the final set of corrected intervals rather than any intermediate prediction. Fig.~\ref{fig:tool_annotation_refinement} shows the annotation-refinement view.

\paragraph{Stage 3: Double Verification and Final Audit.}
Each refined query--annotation pair is independently re-examined by a second annotator who has not seen the first annotator's edits. The two annotators' outputs are compared via temporal Intersection-over-Union (IoU); pairs reaching $\mathrm{IoU} > 0.7$ on \emph{all} corresponding intervals are forwarded to the final benchmark with the primary annotator's annotation. Pairs that fail the IoU threshold are returned for adjudication, in which both annotators jointly review the disputed intervals with the senior annotator until consensus is reached. Only query--annotation pairs passing all three verification stages are included in the final benchmark. The completed annotations are exported as JSON from the annotation tool and merged into the released CoMET-Bench dataset.

\begin{table}[ht]
\centering
\small
\caption{\textbf{Human verification effort.} Stage 1 is averaged over 3,000 candidate queries; Stages 2--3 over 2,789 retained queries. Total time is approximately 1,021 seconds per retained query and excludes training/tool development. Eleven annotators participated; three additionally served as adjudicators.}
\label{tab:annotation_effort_rebuttal}
\begin{tabular}{lrr}
\toprule
\textbf{Stage} & \textbf{Person-hours} & \textbf{Seconds/query} \\
\midrule
Query verification & 85.8 & 103 \\
Annotation refinement & 350.2 & 452 \\
Double verification/adjudication & 354.8 & 458 \\
\midrule
Reported total & $\approx$791 & 1,021 \\
\bottomrule
\end{tabular}
\end{table}

\subsection{Annotation Quality Statistics}
\label{sec:app_quality_stats}

To quantify the rigor of the human verification pipeline (Sec.~\ref{sec:app_human_verification}), Tab.~\ref{tab:quality_stats} reports aggregate intervention statistics across the 600 videos and 2{,}789 final query--annotation pairs. The numbers confirm that MLLM outputs alone are insufficient as ground truth and that human review meaningfully reshapes the released benchmark. At the query stage, only 15.4\% of candidate queries were accepted unchanged: 77.6\% required rewriting (typically to tighten ambiguous wording, fix attribute--query mismatches, or remove spatial-only counting), and 7.0\% were discarded outright. At the annotation stage, 31.7\% of MLLM-proposed intervals were removed as spurious and 26.0\% of the final intervals were newly added by annotators to recover missed instances. Together these numbers indicate that without human refinement the benchmark would substantially under-report event density and over-report condition compliance. In Stage~3, every query was independently re-annotated by a second annotator, and only 6.5\% (182 / 2{,}789) of queries fell below the $\mathrm{tIoU} > 0.7$ agreement threshold and were adjudicated by a senior annotator. The remaining 93.5\% reached direct inter-annotator consensus, supporting the reliability of CoMET-Bench as a benchmark.

\begin{table}[h]
  \centering
  \small
  \setlength{\tabcolsep}{8pt}
  \renewcommand{\arraystretch}{1.15}
  \caption{\textbf{Annotation quality statistics for CoMET-Bench.} Query Verification Stage human intervention on MLLM-generated queries; Annotation Refinement reports interval-level edits on MLLM-proposed annotations; Double Verification reports adjudication outcomes when two independent annotators disagreed on a query.}
  \label{tab:quality_stats}
  \begin{tabular}{l r}
    \toprule
    \textbf{Statistic} & \textbf{Value} \\
    \midrule
    \multicolumn{2}{l}{\textit{Query Verification (3000 $\to$ 2789)}} \\
    \quad Candidate queries generated by MLLM             & 3,000 \\
    \quad Accepted unchanged                              & 462 (15.4\%) \\
    \quad Modified by human verifiers                     & 2,327 (77.6\%) \\
    \quad Discarded                                       & 211 (7.0\%) \\
    \midrule
    \multicolumn{2}{l}{\textit{Annotation Refinement (16247 $\to$ 15327)}} \\
    \quad Candidate intervals proposed by MLLM            & 16,247 \\
    \quad Spurious intervals removed                      & 5,148 (31.7\%) \\
    \quad Missed intervals newly added by annotators      & 4,228 (26.0\%) \\
    \midrule
    \multicolumn{2}{l}{\textit{Double Verification (all 2789 queries re-annotated)}} \\
    \quad Queries requiring third-party adjudication ($\mathrm{tIoU} < 0.7$) & 182 (6.5\%) \\
    \quad Intervals adjudicated within those queries      & 305 \\
    \bottomrule
  \end{tabular}
\end{table}

\subsection{Benchmark Evaluation}
\label{sec:app_benchmark_eval_prompt}

To evaluate baseline MLLMs on CoMET-Bench, we adopt a single, deliberately compact prompt that is shared across all models. The prompt design balances two competing requirements: (i)~it must be sufficiently informative to allow the model to interpret compositional temporal and spatial conditions in the query, and (ii)~it must remain short enough for smaller open-source models to follow without being overwhelmed by lengthy task instructions. Three task-level properties are explicitly enforced. First, the prompt requires the model to identify \emph{every} qualifying interval rather than only the most salient match, directly aligning with the multi-instance nature of our task. Second, it allows the model to return an empty list when no qualifying interval exists, which is essential for fair evaluation on the negative-query subset. Third, the prompt explicitly forbids interval fabrication (\emph{``Do not fabricate intervals''}), which is necessary to prevent models from inflating recall on negative queries by guessing. The output format is a strict JSON array of \texttt{[start\_seconds, end\_seconds]} pairs in integer seconds, which is directly parseable by our evaluation pipeline without post-processing. We use the same prompt for both proprietary and open-source models to ensure a controlled comparison.

\begin{promptbox}{Benchmark Evaluation Prompt}
\small
You are a video analysis assistant. You are given \texttt{\{NUM\_FRAMES\}} frames uniformly sampled from a \texttt{\{duration\_str\}} video.

\vspace{4pt}
\textbf{Query:} \texttt{\{query\_text\}}

\vspace{4pt}
Find every time interval in the video where the query is satisfied. Return all matching intervals, or \texttt{[]} if none exist. \textbf{Do not fabricate intervals.}

\vspace{4pt}
Output ONLY a JSON array of \texttt{[start\_seconds, end\_seconds]} pairs in integer seconds.

\vspace{4pt}
\textbf{Examples:} \texttt{[[10, 20], [45, 60]]} \quad or \quad \texttt{[]}.

\vspace{4pt}
No explanation, no extra text.
\end{promptbox}


\subsection{Evaluation Metrics: Detailed Formulations}
\label{sec:app_metric_formulations}

This section provides the formal definitions of the metrics summarized in Sec.~\ref{evaluation metrics}. Let $\mathcal{Q}$ denote the full query set, partitioned into positive queries $\mathcal{Q}^+$ (with at least one ground-truth interval) and negative queries $\mathcal{Q}^-$ (with zero ground-truth intervals). For a query $q \in \mathcal{Q}$, let $\mathcal{G}_q = \{(s_i^\mathrm{gt}, e_i^\mathrm{gt})\}_{i=1}^{N_q}$ denote the ground-truth interval set with $N_q = |\mathcal{G}_q|$, and let $\mathcal{P}_q = \{(s_j^\mathrm{pred}, e_j^\mathrm{pred})\}_{j=1}^{\hat{N}_q}$ denote the predicted interval set with $\hat{N}_q = |\mathcal{P}_q|$.

\paragraph{Multi-event counting.} We compute three counting metrics over the full query set $\mathcal{Q}$:
\begin{equation}
\mathrm{MAE} = \frac{1}{|\mathcal{Q}|} \sum_{q \in \mathcal{Q}} \big|\hat{N}_q - N_q\big|,
\qquad
\mathrm{OBO} = \frac{1}{|\mathcal{Q}|} \sum_{q \in \mathcal{Q}} \mathbb{1}\big[\,|\hat{N}_q - N_q| \le 1\,\big],
\end{equation}
\begin{equation}
\mathrm{Pearson} = \frac{\sum_{q \in \mathcal{Q}} (\hat{N}_q - \bar{\hat{N}})(N_q - \bar{N})}{\sqrt{\sum_{q \in \mathcal{Q}} (\hat{N}_q - \bar{\hat{N}})^2}\sqrt{\sum_{q \in \mathcal{Q}} (N_q - \bar{N})^2}},
\end{equation}
where $\bar{\hat{N}}$ and $\bar{N}$ are the means of predicted and ground-truth counts. MAE captures average error magnitude, OBO reports the fraction of predictions within a single count of the target (a near-correctness tolerance expected in practical use), and Pearson captures systematic agreement beyond pointwise error~\cite{hu2022transrac-repcount-data, ucfrep-data, countix-data, tsuchiya2026ec-bench}.

\paragraph{Temporal event grounding.} Intuitively, mIoU captures boundary precision, Recall@0.5 captures completeness, and F1@0.5 jointly penalizes over- and under-prediction. Following common practice in temporal grounding \cite{caba2015activitynet, yuan2025momentseeker, gao2017tall-charades-sta-data}, we report the threshold $\theta = 0.5$ in the main paper as a representative operating point. For a single interval pair, the temporal Intersection-over-Union is
\begin{equation}
\mathrm{tIoU}\big((s, e), (s', e')\big) = \frac{\max(0,\, \min(e, e') - \max(s, s'))}{(e - s) + (e' - s') - \max(0,\, \min(e, e') - \max(s, s'))}.
\end{equation}
For each positive query $q \in \mathcal{Q}^+$, we compute three per-query scores and then macro-average them across $\mathcal{Q}^+$. \textbf{mIoU} averages the best-match tIoU over each ground-truth interval:
\begin{equation}
\mathrm{mIoU}_q = \frac{1}{N_q} \sum_{i=1}^{N_q} \max_{j} \mathrm{tIoU}\big((s_j^\mathrm{pred}, e_j^\mathrm{pred}),\, (s_i^\mathrm{gt}, e_i^\mathrm{gt})\big).
\end{equation}
\textbf{Recall@$\theta$} measures the fraction of ground-truth intervals matched at the single event tIoU threshold $\theta$:
\begin{equation}
\mathrm{Recall@}\theta_{\,q} = \frac{1}{N_q} \sum_{i=1}^{N_q} \mathbb{1}\!\left[\,\max_{j} \mathrm{tIoU}\big((s_j^\mathrm{pred}, e_j^\mathrm{pred}),\, (s_i^\mathrm{gt}, e_i^\mathrm{gt})\big) \ge \theta\right].
\end{equation}
\textbf{F1@$\theta$} is computed via greedy one-to-one matching: each prediction is matched to at most one ground-truth interval (in order), and a match is counted as a true positive iff its tIoU exceeds $\theta$. Letting $\mathrm{TP}_q$ denote the number of matched pairs,
\begin{equation}
\mathrm{Precision@}\theta_{\,q} = \frac{\mathrm{TP}_q}{\hat{N}_q},
\quad
\mathrm{F1@}\theta_{\,q} = \frac{2\,\mathrm{Precision@}\theta_{\,q} \cdot \mathrm{Recall@}\theta_{\,q}}{\mathrm{Precision@}\theta_{\,q} + \mathrm{Recall@}\theta_{\,q}}.
\end{equation}

\paragraph{Negative query recognition.} A correct response to a negative query $q \in \mathcal{Q}^-$ is the empty prediction set $\mathcal{P}_q = \emptyset$. The \textbf{Rejection Rate} is the fraction of negative queries handled correctly:
\begin{equation}
\mathrm{RejRate} = \frac{1}{|\mathcal{Q}^-|} \sum_{q \in \mathcal{Q}^-} \mathbb{1}\big[\,\hat{N}_q = 0\,\big].
\end{equation}
However, RejRate alone is trivially gameable: a model that always emits an empty prediction reaches $\mathrm{RejRate} = 1$ at the cost of zero recall on positive queries. To prevent this, we additionally define the \textbf{Positive Coverage} as the fraction of positive queries on which the model returns at least one prediction,
\begin{equation}
\mathrm{PosCov} = \frac{1}{|\mathcal{Q}^+|} \sum_{q \in \mathcal{Q}^+} \mathbb{1}\big[\,\hat{N}_q > 0\,\big],
\end{equation}
and report \textbf{Rej.-F1} as the harmonic mean of the two,
\begin{equation}
\mathrm{Rej.-F1} = \frac{2 \cdot \mathrm{RejRate} \cdot \mathrm{PosCov}}{\mathrm{RejRate} + \mathrm{PosCov}}.
\end{equation}
Rej.-F1 takes a high value only when the model both correctly rejects negatives and remains willing to commit predictions on positives, ruling out the lazy ``always-empty'' strategy. As a complementary readout of hallucination severity, we also report the \textbf{False Positive Rate (FPR)}, defined as the fraction of negative queries on which the model emits any non-empty prediction. While FPR is the direct complement of RejRate ($\mathrm{FPR} = 1 - \mathrm{RejRate}$), reporting it alongside Rej.-F1 makes the magnitude of hallucination on negative queries directly visible in the result table.


\section{Additional Details of Methodology}
\label{appendix_method}

This appendix complements Sec.~\ref{method} by providing the formal definitions and algorithmic details that were omitted in the main paper for brevity. We elaborate on three components: the temporal-window parsing performed by the Planner Agent (Sec.~\ref{sec:app_preseg}), the edge weight formulation used to construct the hierarchical graph (Sec.~\ref{sec:app_graph_edges}), and the iterative verification algorithm with online deduplication that drives Steps~3 and~4 (Sec.~\ref{sec:app_algorithm}). All notation is consistent with the main text.

\subsection{Adaptive Video Planning (Step 1)}
\label{sec:app_preseg}

\paragraph{Pre-segmentation.} When a query specifies an explicit temporal window (e.g., \emph{``Between 00:16:00 and 00:34:00...''}), processing the entire video introduces unnecessary computational overhead and risks hallucination from out-of-bound segments. The \emph{Planner Agent} parses such queries to extract the absolute boundary timestamps $T_\mathrm{start}$ and $T_\mathrm{end}$, and trims the original video $\mathcal{V}_\mathrm{full}$ into a pre-segmented video:
\begin{equation}
  \mathcal{V}_\mathrm{trim} = \{f_t \in \mathcal{V}_\mathrm{full} \mid T_\mathrm{start} \le t \le T_\mathrm{end}\},
\end{equation}
where $f_t$ denotes the video frame at timestamp $t$. The trimmed clip $\mathcal{V}_\mathrm{trim}$ is then passed to Step~2, ensuring that all subsequent graph construction and reasoning are strictly bounded within the user-specified temporal scope. For queries without explicit time bounds, $T_\mathrm{start}$ and $T_\mathrm{end}$ default to $0$ and the full video duration, respectively.

\paragraph{Hyperparameter Adaptive Configuration.}
The granularity of the hierarchical graph strongly affects downstream verification: too coarse a segmentation collapses distinct events into a single node, while too fine a segmentation fragments a single event across many spurious nodes. Since the optimal granularity depends on both the visual dynamics of the video and the temporal scope of the query, no single static setting generalizes across the benchmark. We therefore pre-define a small library of \emph{configuration profiles}, where each profile is a complete, internally consistent set of hyperparameters that controls graph construction. A profile specifies the frame sampling rate used to extract DINOv2 and RAFT features, the event-level boundary threshold $\tau_\mathrm{e}$ that governs event-node segmentation, the PELT penalty that governs action-node segmentation, and the edge parameters $\sigma$ and $\tau$ from Eq.~\ref{eq:edge_weight} that control graph traversal. Each profile is tuned for a representative regime, namely a \emph{coarse} profile for short videos with a broad temporal scope, a \emph{balanced} profile for typical long-form videos, and a \emph{fine} profile for fast-paced or narrow-window queries. Given the query and a brief description of the trimmed video $\mathcal{V}_\mathrm{trim}$, the \emph{Planner Agent} selects the most suitable profile, which is then propagated to all downstream operations in Step~2. By delegating profile-level selection to the agent rather than asking it to set numerical hyperparameters directly, the framework retains adaptivity while avoiding the brittleness of free-form numerical generation by an MLLM.

\subsection{Temporal Graph Building (Step 2)}
\label{sec:app_graph_edges}

This section details the change-point detection procedure used to construct the hierarchical graph in Step~2 (Sec.~\ref{sec:graph_building}) and the edge weight formulation that quantifies inter-node affinity.

\paragraph{Event-level segmentation.}
Given the pre-segmented video $\mathcal{V}_\mathrm{trim}$, we sample frames at a fixed rate (1\,fps in our default configuration) and extract per-frame semantic features $\{\mathbf{x}_t\}_{t=1}^{T}$ using a frozen DINOv2 encoder~\cite{simeoni2025dinov3}, where $T$ is the total number of sampled frames. We compute the inter-frame cosine similarity sequence
\begin{equation}
  s^\mathrm{e}_t \;=\; \frac{\mathbf{x}_t^\top \mathbf{x}_{t-1}}{\|\mathbf{x}_t\|\,\|\mathbf{x}_{t-1}\|}, \qquad t = 2, \dots, T,
\end{equation}
and apply wavelet smoothing (Daubechies-4 with soft thresholding) to suppress high-frequency noise while preserving genuine scene transitions. Event-level boundaries are then identified as the local minima of the smoothed curve that fall below an adaptive threshold $\tau_\mathrm{e}$ (set by the Planner Agent), since semantically coherent events correspond to plateaus of high inter-frame similarity, separated by sharp similarity drops at scene changes. Each contiguous segment between consecutive boundaries forms an Event Node $v_i \in \mathcal{V}_E$ with timestamp interval $[t_i^\mathrm{start}, t_i^\mathrm{end}]$.

\paragraph{Action-level segmentation.}
For each event node $v_i$ retained by the Filter Agent, we apply a second segmentation pass that operates on motion rather than appearance. We extract dense optical flow $\{\mathbf{f}_t\}_{t=1}^{T_i-1}$ between consecutive frames within $v_i$ using a frozen RAFT encoder~\cite{teed2020raft}, where $T_i$ is the number of frames in $v_i$. We then compute the per-frame flow magnitude
\begin{equation}
  g_t \;=\; \|\mathbf{f}_t\|_2,
\end{equation}
which is large at frames containing strong motion (likely action boundaries) and small within motion-homogeneous regions. To convert this signal into a similarity-style curve whose local minima mark transitions, we negate and rescale it to $[0, 1]$ via min--max normalization:
\begin{equation}
  s^\mathrm{a}_t \;=\; \frac{g_\mathrm{max} - g_t}{g_\mathrm{max} - g_\mathrm{min} + \epsilon},
  \qquad
  g_\mathrm{max} = \max_{1 \le t \le T_i-1} g_t,\ \ g_\mathrm{min} = \min_{1 \le t \le T_i-1} g_t,
\end{equation}
where $\epsilon$ is a small constant for numerical stability. After this transformation, frames with low motion correspond to $s^\mathrm{a}_t \approx 1$ (motion-homogeneous regions) and frames with strong motion correspond to $s^\mathrm{a}_t \approx 0$ (likely action transitions). We further smooth $s^\mathrm{a}$ using wavelet denoising with hard thresholding (which preserves abrupt motion changes better than soft thresholding) and apply the PELT change-point detection algorithm~\cite{killick2012change_point_detection} with an RBF kernel cost to recover action-level boundaries. The resulting fine-grained segments form the Action Nodes $\mathcal{V}_A$ within each filtered event.

\paragraph{Edge weight formulation.}
Both the Event Graph $\mathcal{G}_E$ and the Action Graph $\mathcal{G}_A$ contain \emph{direct neighbor edges} (between temporally adjacent nodes) and \emph{indirect neighbor edges} (between non-adjacent nodes within a bounded hop distance), as introduced in Sec.~\ref{sec:graph_building}. To quantify the strength of each edge---and to provide a principled criterion for traversing the graph during iterative verification (Sec.~\ref{iterative graph-based verification})---we assign a scalar weight that combines semantic similarity with a temporal-decay penalty.

For any two nodes $v_i, v_j \in \mathcal{V}$, we extract segment-level visual representations $\mathbf{z}_i, \mathbf{z}_j \in \mathbb{R}^d$ using the frozen video embedding model~\cite{bai2025qwen3vl} and define
\begin{equation}
  \label{eq:edge_weight}
  w(e_{i,j}) \;=\; \underbrace{\frac{\mathbf{z}_i^\top \mathbf{z}_j}{\|\mathbf{z}_i\|\,\|\mathbf{z}_j\|}}_{\text{semantic similarity}} \cdot \underbrace{\exp\!\left(-\frac{|i - j|}{\sigma}\right)}_{\text{temporal decay}},
\end{equation}
where $|i - j|$ denotes the discrete hop distance between $v_i$ and $v_j$ along the temporal sequence, and $\sigma > 0$ is a scalar controlling how rapidly the weight decays with temporal distance (set adaptively by the Planner Agent). This formulation favors connections between nodes that are simultaneously semantically related and temporally proximate, suppressing spurious long-range similarities. During the iterative verification of Step~3, only edges satisfying $w(e_{i,j}) > \tau$ are traversed when constructing the perception context $\mathcal{C}(v_i)$, where $\tau$ is the edge threshold introduced in the main text.

\subsection{Iterative Verification with Online Deduplication (Step 3)}
\label{sec:app_algorithm}

This section formalizes the iterative refinement loop introduced in Sec.~\ref{iterative graph-based verification} and the online update mechanism (i.e., online deduplication) that populates the Global Memory Bank $\mathcal{M}$ described in Sec.~\ref{global memory bank}. It takes as input the fine-grained Action Graph $\mathcal{G}_A = (\mathcal{V}_A, \mathcal{E}_A)$ constructed in Step~2 and the natural-language query $Q$, and returns a populated memory bank $\mathcal{M}$ containing every verified, deduplicated event instance. Algorithm~\ref{alg:verification} provides the complete pseudocode. The algorithm maintains an unverified node set $\mathcal{U} \subseteq \mathcal{V}_A$ (initialized to $\mathcal{V}_A$) and an iteration counter $\mathit{iter}$. Each iteration consists of three operations.

\begin{algorithm}[h]
  \caption{Iterative Graph-based Verification with Online Deduplication}
  \label{alg:verification}
  \textbf{Input:} Action Graph $\mathcal{G}_A = (\mathcal{V}_A, \mathcal{E}_A)$, query $Q$, edge threshold $\tau$, IoU threshold $\gamma$, max hops $K$, max iterations $N_\mathrm{max}$ \\
  \textbf{Output:} Populated Global Memory Bank $\mathcal{M}$
  \begin{algorithmic}[1]
    \STATE Initialize $\mathcal{M} \leftarrow \emptyset$,\ \ $\mathcal{U} \leftarrow \mathcal{V}_A$,\ \ $\mathit{iter} \leftarrow 0$
    \WHILE{$\mathcal{U} \neq \emptyset$ \AND $\mathit{iter} < N_\mathrm{max}$}
    \STATE Sample candidate node $v_i \sim \mathcal{U}$
    \STATE \hl{\emph{\# (1) Node Reduction \& Combination}}
    \STATE $\mathcal{C}(v_i) \leftarrow \{v_i\} \cup \mathcal{N}_K(v_i;\, \tau)$ \COMMENT{$K$-hop neighbors with edge weight $w > \tau$}
    \STATE \hl{\emph{\# (2) Verifier Agent inspection}}
    \STATE $m \leftarrow \mathrm{VerifierAgent}\bigl(\mathcal{C}(v_i),\, Q\bigr)$
    \IF{$m$ is valid}
    \STATE Parse $m = \langle \mathcal{S},\, (t^\mathrm{start}, t^\mathrm{end}),\, c \rangle$
    \STATE \hl{\emph{\# (3) Online deduplication via temporal NMS}}
    \STATE $\mathit{is\_dup} \leftarrow \textbf{False}$
    \FOR{each $m' = \langle \mathcal{S}',\, (t'^\mathrm{start}, t'^\mathrm{end}),\, c' \rangle \in \mathcal{M}$}
    \IF{$\mathrm{IoU}\bigl((t^\mathrm{start}, t^\mathrm{end}),\, (t'^\mathrm{start}, t'^\mathrm{end})\bigr) > \gamma$}
    \STATE $\mathit{is\_dup} \leftarrow \textbf{True}$
    \IF{$c > c'$}
    \STATE $\mathcal{M} \leftarrow (\mathcal{M} \setminus \{m'\}) \cup \{m\}$ \COMMENT{Replace lower-confidence duplicate}
    \ENDIF
    \ENDIF
    \ENDFOR
    \IF{\NOT $\mathit{is\_dup}$}
    \STATE $\mathcal{M} \leftarrow \mathcal{M} \cup \{m\}$ \COMMENT{Append novel event}
    \ENDIF
    \ENDIF
    \STATE \hl{\emph{\# Mark covered nodes to avoid redundant search}}
    \STATE $\mathcal{U} \leftarrow \mathcal{U} \setminus \mathcal{C}(v_i)$
    \STATE $\mathit{iter} \leftarrow \mathit{iter} + 1$
    \ENDWHILE
    \RETURN $\mathcal{M}$
  \end{algorithmic}
\end{algorithm}

\paragraph{(1) Node Reduction and Combination.}
A candidate action node $v_i$ is sampled from $\mathcal{U}$. To enlarge the limited temporal receptive field of an isolated action, we form a perception context $\mathcal{C}(v_i)$ by unioning $v_i$ with its $K$-hop neighborhood, where neighbor traversal is restricted to edges whose weight $w(e_{i,j}) > \tau$ (with $w$ defined in ~\ref{eq:edge_weight} and $\tau$ being the edge threshold from the main text):
\begin{equation}
  \mathcal{C}(v_i) = \{v_i\} \cup \mathcal{N}_K(v_i;\, \tau),
\end{equation}
where $\mathcal{N}_K(v_i;\, \tau)$ denotes the set of nodes reachable from $v_i$ within $K$ hops along edges satisfying $w > \tau$. This step realizes the Node Reduction and Combination operation introduced in Sec.~\ref{iterative graph-based verification}.

\paragraph{(2) Verifier Agent inspection.}
Frames sampled from the perception context $\mathcal{C}(v_i)$ are passed, together with the query $Q$, to the Verifier Agent. The agent decides whether the candidate action node $v_i$ satisfies all spatial and temporal conditions specified by $Q$. If verified, the agent emits a memory tuple
\begin{equation}
  m = \langle \mathcal{S},\, (t^\mathrm{start}, t^\mathrm{end}),\, c \rangle,
\end{equation}
following the definition in Sec.~\ref{global memory bank}: $\mathcal{S}$ is a textual summary describing the verified event and its salient entities, $(t^\mathrm{start}, t^\mathrm{end})$ are the timestamps inherited from $v_i$, and $c \in [0, 1]$ is the Verifier's self-reported confidence. If the agent rejects $v_i$, no tuple is produced.

\paragraph{(3) Online deduplication via temporal NMS.}
A newly verified tuple $m$ may correspond to the same underlying event as an existing entry $m' \in \mathcal{M}$, since multiple candidate nodes within the same event region can independently trigger verification. To prevent redundant detections, we apply an online temporal Non-Maximum Suppression: $m$ is compared against every existing $m' = \langle \mathcal{S}',\, (t'^{\mathrm{start}}, t'^{\mathrm{end}}),\, c' \rangle \in \mathcal{M}$ via temporal Intersection-over-Union (IoU); if any overlap exceeds the threshold $\gamma$, the framework retains only the tuple with the higher confidence $c$. If no existing entry overlaps, $m$ is appended as a novel event. After processing $v_i$, all nodes in $\mathcal{C}(v_i)$ are removed from $\mathcal{U}$ to avoid redundant verification of nodes already covered by the perception context. The loop terminates when either $\mathcal{U}$ is exhausted or the iteration count reaches $N_\mathrm{max}$. The resulting memory bank $\mathcal{M}$ is then consumed by the Aggregator Agent in Step~4 to produce the final grounded intervals and exact count.

\subsection{Global Memory Bank Mechanism}
\label{sec:app_global_memory_bank}

This section provides the detailed formulation of the Global Memory Bank introduced in Sec.~\ref{global memory bank}, including its tuple semantics, the online deduplication policy, and the cross-moment aggregation performed by the Aggregator Agent in Step~4.

\paragraph{Memory entries.}
The Global Memory Bank, denoted as $\mathcal{M}$, is a persistent data structure that supports global perception across the entire long-form video and overcomes the limitations of localized processing. For every node verified by the Verifier Agent in Step~3, the framework emits an entry
\begin{equation}
m = \langle \mathcal{S},\, (t^\mathrm{start}, t^\mathrm{end}),\, c \rangle,
\end{equation}
where $\mathcal{S}$ is a concise text summary describing the visual event and the salient features of the involved entities, $(t^\mathrm{start}, t^\mathrm{end})$ are the grounding timestamps inherited from the fine-grained action node, and $c \in [0, 1]$ is the agent's self-reported confidence score. The text summary $\mathcal{S}$ serves a dual purpose: it both records what was verified and provides the cross-moment evidence consumed by the Aggregator Agent in Step~4.

\paragraph{Online deduplication.}
Because neighboring action nodes covering the same underlying event may independently trigger verification, naively appending every verified entry would inflate $\mathcal{M}$ with near-duplicate records. The memory bank employs an \emph{online deduplication} mechanism (formalized in Algorithm~\ref{alg:verification}): a new entry is appended as a novel event if its temporal IoU with all existing entries falls below threshold $\gamma$; otherwise only the higher-confidence entry is retained. The iterative search continues until the maximum iteration is reached or all candidate nodes in $\mathcal{G}_A$ have been processed, populating $\mathcal{M}$ with a comprehensive, deduplicated log of all condition-compliant events.

\paragraph{Cross-moment aggregation.}
In Step~4, the \emph{Aggregator Agent} consolidates the populated memory bank $\mathcal{M}$ into the final answer: an exact count $\hat{N}$ of the validated instances and their corresponding grounded intervals $\{(t_j^\mathrm{start}, t_j^\mathrm{end})\}_{j=1}^{\hat{N}}$. Beyond simple enumeration, the Aggregator performs \emph{cross-moment reasoning}: by cross-referencing the descriptive summaries $\mathcal{S}$ across all stored entries, it determines whether temporally disjoint events involve the same underlying entities, resolving identity consistency for queries that refer to specific individuals or objects (e.g., \emph{``the referee who first issued a yellow card''}). This step is what allows CoMET-Agent to handle Identity-attribute queries that single-pass inference cannot, since identity verification requires comparing evidence collected at different timestamps. The resulting answer, $(\hat{N}, \{(t_j^\mathrm{start}, t_j^\mathrm{end})\}_{j=1}^{\hat{N}})$, is the structured output of the search-and-aggregate pipeline.

\section{Implementation Details}
\label{implementation details}

This section reports the key implementation choices used in our experiments, focusing only on the parameters that appear in the main paper or directly affect the reported results. Code and configuration files will be released upon publication.

\paragraph{Model choices.}
CoMET-Agent is fully training-free and assembled from frozen, off-the-shelf components. The four agent roles (\emph{Planner}, \emph{Filter}, \emph{Verifier}, \emph{Aggregator}) are all instantiated by the same MLLM, \textbf{Qwen3-VL-8B-Instruct}~\cite{bai2025qwen3vl}, in zero-shot mode without any task-specific fine-tuning. Edge weights in the hierarchical graph (Eq.~\ref{eq:edge_weight}) are computed using \textbf{Qwen3-VL-Embedding-8B}~\cite{bai2025qwen3vl} as the segment-level video encoder. For the visual feature extractors used in change-point detection (Sec.~\ref{sec:app_graph_edges}), we adopt frozen \textbf{DINOv2}~\cite{simeoni2025dinov3} (ViT-L/14 backbone) for event-level appearance features and frozen \textbf{RAFT}~\cite{teed2020raft} for action-level optical flow. All experiments run on a single NVIDIA A100~(80\,GB) GPU. We do the evaluation of other methods on multiple NVIDIA 4$\times$A40 GPUs.

\paragraph{Hyperparameters of iterative verification.}
The iterative loop in Step~3 is bounded by $N_\mathrm{max} = 3$ iterations for both node reduction (Step~3.1) and verification (Step~3.2), with early termination when no new candidates are accepted. Online deduplication uses a temporal IoU threshold of $\gamma = 0.5$.

\paragraph{Configuration profiles for the hierarchical graph.}
As described in Sec.~\ref{sec:app_preseg}, the Planner Agent selects one of three pre-defined configuration profiles (\emph{Short}, \emph{Medium}, \emph{Long}) based on the query and the trimmed video. Each profile is a complete, internally consistent set of hyperparameters tuned for a specific duration regime; the concrete values are summarized in Tab.~\ref{tab:profiles}. The \emph{Short} profile (videos under 10 minutes) uses denser sampling and stricter thresholds to capture fast, fine-grained transitions; the \emph{Medium} profile (10--30 minutes) balances granularity and computational cost; the \emph{Long} profile (over 30 minutes) uses coarser sampling and looser thresholds to keep the graph manageable while preserving coverage of long-range dependencies.

\begin{table}[ht]
\centering
\small
\renewcommand{\arraystretch}{0.8}
\caption{Configuration profiles selected by the Planner Agent in Step~1, applied to event- and action-level graph construction in Step~2 and to graph traversal in Step~3.}
\label{tab:profiles}
\begin{tabular}{l c c c}
\toprule
\textbf{Parameter} & \textbf{Short} & \textbf{Medium} & \textbf{Long} \\
 & ($<$10\,min) & (10--30\,min) & ($>$30\,min) \\
\midrule
\multicolumn{4}{l}{\emph{Event-level segmentation}} \\
\quad Frame sampling rate $\rho_E$ (fps)             & 2.0  & 1.0  & 0.5  \\
\quad Boundary threshold $\tau_\mathrm{e}$           & 0.92 & 0.88 & 0.82 \\
\midrule
\multicolumn{4}{l}{\emph{Action-level segmentation}} \\
\quad Frame sampling rate $\rho_A$ (fps)             & 2.0  & 2.0  & 1.0  \\
\quad PELT penalty $\beta$                           & 0.30 & 0.50 & 0.70 \\
\midrule
\multicolumn{4}{l}{\emph{Edge construction (Eq.~\ref{eq:edge_weight}) and graph traversal}} \\
\quad Temporal-decay scale $\sigma$                  & 2.0  & 3.0  & 5.0  \\
\quad Edge traversal threshold $\tau$                & 0.85 & 0.82 & 0.78 \\
\midrule
\multicolumn{4}{l}{\emph{Iterative verification budgets}} \\
\quad \textsc{max\_coarse\_events}                   & 3    & 4    & 6    \\
\quad \textsc{max\_action\_nodes}                    & 4    & 6    & 8    \\
\bottomrule
\end{tabular}
\end{table}

\paragraph{Baselines.}
For the open-source MLLM and agent-based baselines reported in Tab.~\ref{tab:main_results}, we use the official checkpoints released by the authors and evaluate them with the prompt in Sec.~\ref{sec:app_benchmark_eval_prompt} under uniform frame sampling at the model's default frame budget. Proprietary models (GPT-5, Gemini~2.5~Pro, Gemini~3~Flash) are queried through their public APIs with the same prompt and a fixed sampling rate of 128 uniformly extracted frames per video to control inference cost.

\subsection{Details of Agent-based Method Benchmarking}
\label{agent method benchmarking}

In Tab.~\ref{tab:main_results}, two agent-based baselines, Vgent~\cite{shen2025vgent} and T$^*$~\cite{ye2025T*}, are reported as ``Cannot Follow Instruction.'' We elaborate on this status below to clarify that it is a structural property of their pipelines rather than an artifact of any specific run.

The benchmark evaluation prompt (Appendix~\ref{sec:app_benchmark_eval_prompt}) requires \textbf{(i)~exhaustive enumeration} of every qualifying interval rather than a single best match, \textbf{(ii)~explicit rejection} via an empty list \texttt{[]} when no qualifying interval exists, and \textbf{(iii)~strict JSON output} of \texttt{[start\_seconds, end\_seconds]} pairs.
These requirements expose two structural mismatches with existing agent pipelines.
\textbf{(1) Task-paradigm mismatch.}
Both Vgent and T$^*$ are originally designed for video question answering (VideoQA), where the terminal output is natural-language answer text rather than temporal intervals. Repurposing such pipelines for compositional temporal grounding requires their internal reasoning steps, which are optimized for producing and verifying answer phrases, to instead emit, enumerate, and verify timestamp pairs, which is a paradigm-level shift that cannot be achieved by prompt edits alone.
\textbf{(2) Pipeline-format rigidity.}
Each pipeline is composed of multiple fixed steps (e.g., planner, retriever, answerer), each of which expects and emits step-specific intermediate formats. Modifying the final output format propagates inconsistencies upstream, since the answerer's expected input is shaped by the retriever's fixed output schema. Despite extensive prompt adaptations on top of the official code, both pipelines consistently produced empty or unparseable outputs on CoMET-Bench, and we accordingly mark them as ``Cannot Follow Instruction.'' We view this as evidence that the conditional multi-event grounding setting demands more general agent compositions than prior question-answering pipelines provide.

\section{Additional Experiments}
\label{additional experiments}

\subsection{Additional Analysis on CoMET-Bench}
\label{additional analysis benchmark}

We provide three complementary analyses that surface patterns hidden by the headline numbers in Tab.~\ref{tab:main_results}.

\paragraph{Scaling improves counting but hurts grounding.}
Two open-source families exhibit the same counter-intuitive divergence under parameter scaling. The Qwen3-VL 8.3B$\,\to\,$30B comparison improves counting (MAE 4.8$\,\to\,$3.6) yet \emph{degrades} grounding (F1@0.5 3.4\%$\,\to\,$3.1\%) and Rej.-F1 (52.1$\,\to\,$47.2). The InternVL3.5 8B$\,\to\,$38B comparison shows the same dissociation: F1@0.5 marginally rises (1.1\%$\,\to\,$1.3\%) but Rej.-F1 collapses (64.7$\,\to\,$40.3) while MAE actually worsens (3.6$\,\to\,$4.0). This consistent pattern across families suggests that simple parameter scaling does not translate into the structured search needed for precise temporal grounding, motivating the dual-axis design of our evaluation protocol.

\paragraph{Counting calibration depends on reasoning capacity more than visual capacity.}
Pearson correlation ranks differently from grounding metrics: the top scores come from VideoARM (14.5), Ours\,(GPT-5) (13.7), InternVL3.5 38B (13.4), and Ours\,(Gemini~3~Flash) (12.0). With the partial exception of InternVL3.5 38B, the top of this ranking is dominated by systems with strong reasoning components (GPT-4.1+o3, GPT-5, Gemini~3), whereas top grounding scores instead follow visual grounding capacity. This suggests count calibration is closer to a reasoning task (``how many qualifying instances did I see?''), while grounding accuracy depends jointly on visual perception and structured search. The dissociation between Pearson and F1@0.5 within the same model further supports that the two axes capture distinct capabilities.

\begin{figure}[h]
  \centering
  \includegraphics[width=0.7\linewidth]{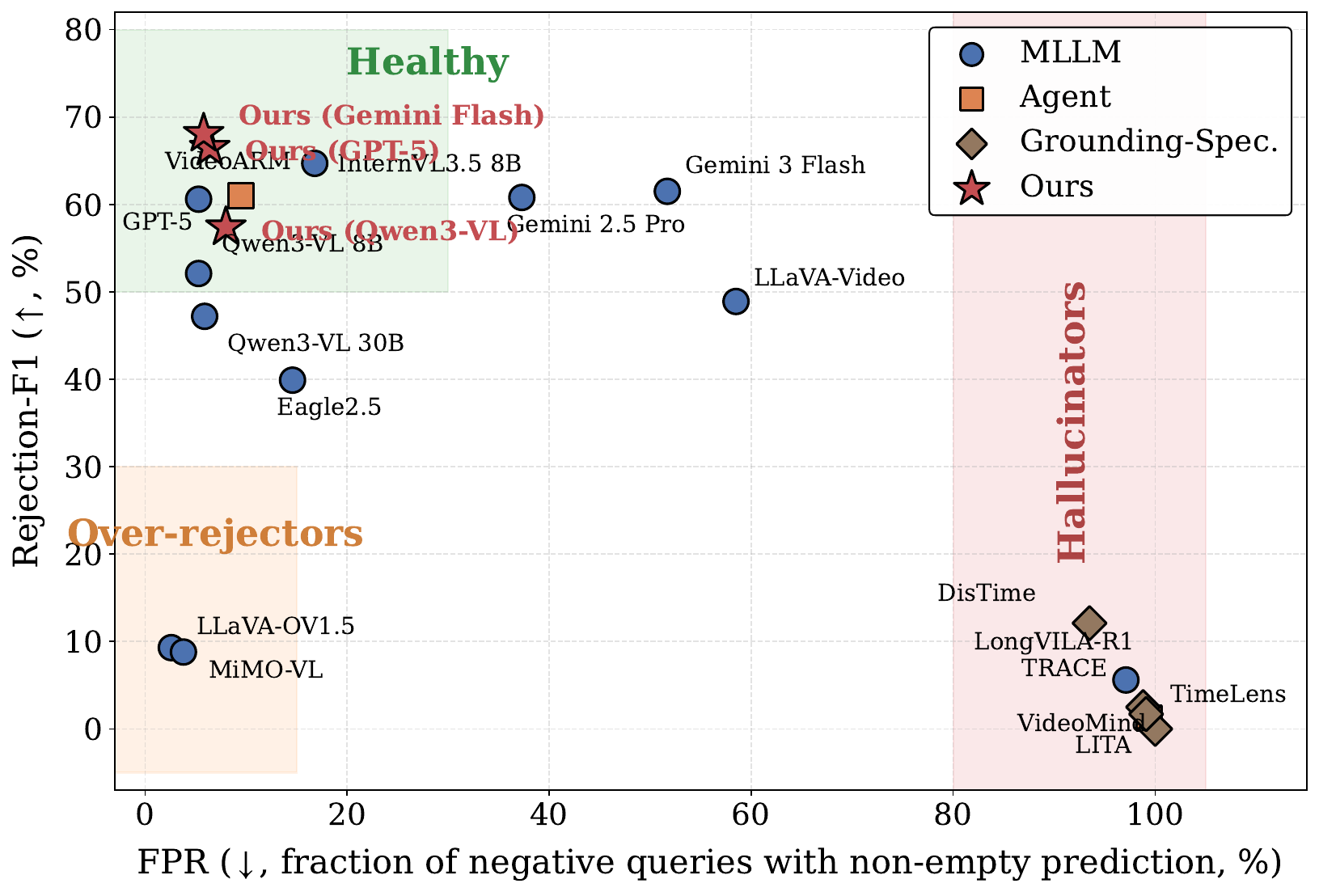}
  \caption{\textbf{Two failure modes on negative queries.} Each point is a method on Tab.~\ref{tab:main_results}; colors denote method category. \emph{Hallucinators} (bottom-right) predict intervals on nearly every negative query, while \emph{over-rejectors} (bottom-left) reject nearly every query, including positives. Only methods in the top-left region achieve both low hallucination and high positive coverage.}
  \label{fig:negative_taxonomy}
\end{figure}

\paragraph{Negative queries split methods into two failure modes.}
Fig.~\ref{fig:negative_taxonomy} visualizes the bimodal pattern on negative queries. \textbf{Hallucinators} predict non-empty intervals on $\geq$93\% of negative queries, including all four grounding-specialized models (TRACE 98.8\%, LITA 99.1\%, DisTime 93.5\%, and TimeLens 100\%, the latter rejecting no negative query at all) and VideoMind (99.4\%). \textbf{Over-rejectors} achieve very low FPR by also rejecting positive queries: LLaVA-OV1.5 (FPR 2.6\%, Rej.-F1 9.3\%) and MiMO-VL (FPR 3.8\%, Rej.-F1 8.8\%) return empty for nearly every query regardless of ground truth, an outcome that the standard rejection rate would falsely reward. Only methods in the middle of this spectrum, including GPT-5 and our CoMET-Agent variants (Rej.-F1 57.4--68.1), achieve both low hallucination and high coverage. This bimodal split is exactly what our proposed \textbf{Rejection-F1} metric is designed to detect.

\subsection{CoMET-Agent's Performance on Existing Benchmarks}
\label{method performance}

To verify that CoMET-Agent generalizes beyond CoMET-Bench, we evaluate it on two standard temporal-grounding benchmarks: Charades-STA~\cite{gao2017tall-charades-sta-data} and ActivityNet-Captions~\cite{activitynet-captions}. Both benchmarks ask the model to localize a single event described in natural language; we follow the standard protocol and report Recall@0.5 (R@0.5) and mIoU. We instantiate CoMET-Agent with the open-source backbone Qwen2.5-VL-7B and compare against representative video grounding methods. Tab.~\ref{tab:cross_bench} reports the results.

CoMET-Agent achieves the strongest zero-shot performance on Charades-STA (R@0.5 56.2\%, mIoU 50.1\%), surpassing all zero-shot baselines including DisTime (52.0\% / 47.4\%) and TimeMarker (51.9\% / 48.4\%), and approaching the fully-supervised TRACE (61.7\% R@0.5). Compared to HiTeA, which uses the same Qwen2.5-VL backbone, CoMET-Agent improves R@0.5 by 7.7\% on Charades-STA and 9.8\% on ActivityNet-Captions, isolating the gain to the framework rather than the backbone. On ActivityNet-Captions, our R@0.5 (40.9\%) is below TimeMarker (50.7\%), which uses a larger 8B backbone with grounding-specific pretraining; nonetheless, CoMET-Agent remains the strongest zero-shot Qwen2.5-VL-based system. These results confirm that the search-and-aggregate framework, although designed for compositional multi-event grounding on long videos, also transfers effectively to classical single-event grounding on shorter clips.

\begin{table}[h]
  \centering
  \small
  \setlength{\tabcolsep}{6pt}
  \renewcommand{\arraystretch}{1.05}
  \caption{\textbf{Comparison with representative video grounding methods on Charades-STA and ActivityNet-Captions.} We report R@0.5 and mIoU under each benchmark's official protocol. \emph{Sup}: supervision regime, with FS = fully supervised, ZS = zero-shot.}
  \label{tab:cross_bench}
  \begin{tabular}{l c cc cc}
    \toprule
    \multirow{2}{*}{\textbf{Method}} & \multirow{2}{*}{\textbf{Sup}}
    & \multicolumn{2}{c}{\textbf{Charades-STA}}
    & \multicolumn{2}{c}{\textbf{ActivityNet-Captions}} \\
    \cmidrule(lr){3-4} \cmidrule(lr){5-6}
    & & R@0.5$\uparrow$ & mIoU$\uparrow$
      & R@0.5$\uparrow$ & mIoU$\uparrow$ \\
    \midrule
    VTimeLLM~\cite{huang2024vtimellm}    & FS & 34.3  & 34.6  & 29.5  & 31.4  \\
    TRACE~\cite{guo2024trace}& FS & 61.7  & -  & 37.7  & 39.0  \\
    \midrule
    Momentor~\cite{qian2024momentor}                     & ZS & 26.6  & 28.5  & 23.0  & 29.3  \\
    ChatVTG~\cite{qu2024chatvtg}                      & ZS & 33.0  & 34.9  & 22.5  & 27.2  \\
    TimeChat~\cite{ren2024timechat}      & ZS & 32.2  & --    & --    & --    \\
    InternVideo2.5~\cite{internvideo2.5}               & ZS & 43.3  & 41.7  & --    & --    \\
    TimeMarker~\cite{chen2024timemarker}                   & ZS & 51.9  & 48.4  & 50.7  & 49.5  \\
    DisTime~\cite{zeng2025distime} (Llava-OV)       & ZS & 52.0  & 47.4  & 34.3  & 37.1  \\
    HiTeA~\cite{hitea} (Qwen2.5-VL)                       & ZS & 48.5  & 46.3  & 31.1  & 37.9  \\
    \midrule
    \rowhl
    \textbf{CoMET-Agent (\underline{Qwen2.5-VL})}    & ZS & 56.2   & 50.1   & 40.9   & 41.6   \\
    \bottomrule
  \end{tabular}
\end{table}

\subsection{Method Ablations}
\label{method ablations}

We disable one component at a time and report Pearson correlation (counting calibration), grounding F1@0.5, and Rejection-F1 (Rej.-F1) on CoMET-Bench. All variants are instantiated with the Qwen3-VL-7B backbone for a controlled comparison. Tab.~\ref{tab:ablation} summarizes the results.

\noindent\textbf{Findings.}
\textbf{(1) Every component contributes:} all four ablated variants fall between the raw Qwen3-VL baseline (Pearson 4.8\%, F1@0.5 3.4\%, Rej.-F1 52.1\%) and the full CoMET-Agent (11.0\% / 8.8\% / 57.4\%), confirming that each module addresses a distinct failure mode rather than overlapping with the others.
\textbf{(2) Global Memory Bank is the most impactful:} removing it drops Pearson to 6.0\% and F1@0.5 to 4.0\%, the largest absolute drops on both metrics, since the agent loses cross-moment evidence and degenerates to local-only verification.
\textbf{(3) Iterative Verification is critical for negative-query handling:} disabling it collapses Rej.-F1 to 52.1\%, identical to the raw baseline, indicating that without iterative context expansion the verifier cannot reliably reject hallucinated events.
\textbf{(4) Temporal Graph and Adaptive HP provide complementary smaller gains:} removing the hierarchical graph (flat segmentation) drops F1@0.5 by 2.3\% and Rej.-F1 by 2.5\%, mostly hurting long-video performance, while removing adaptive HP configuration causes the smallest drop (-1.3\% F1@0.5), as static profiles still cover the typical 30-minute regime adequately.

\begin{table}[h]
  \centering
  \small
  \setlength{\tabcolsep}{6pt}
  \renewcommand{\arraystretch}{1.1}
  \caption{\textbf{Ablation study of CoMET-Agent on CoMET-Bench (Qwen3-VL 8B backbone).} We disable one component at a time and report Pearson correlation, grounding F1@0.5, and Rejection-F1.}
  \label{tab:ablation}
  \begin{tabular}{l ccc}
    \toprule
    \textbf{Variant} & \textbf{Pearson~$\uparrow$} & \textbf{F1@0.5 $\uparrow$} & \textbf{Rej.-F1 $\uparrow$} \\
    \midrule
    Qwen3-VL 7B (baseline)                     & 4.8   & 3.4   & 52.1   \\
    \midrule
    \quad w/o Adaptive HP Configuration        & 9.2   & 7.5   & 56.3   \\
    \quad w/o Temporal Graph (flat segm.)      & 7.7   & 6.5   & 54.9   \\
    \quad w/o Iterative Verification           & 6.2   & 4.2   & 52.1   \\
    \quad w/o Global Memory Bank               & 6.0   & 4.0   & 53.8   \\
    \midrule
    \rowhl
    \textbf{CoMET-Agent (Qwen3-VL 8B)}         & 11.0  & 8.8   & 57.4   \\
    \bottomrule
  \end{tabular}
\end{table}

\begin{table}[ht]
\centering
\small
\caption{\textbf{Completed latency comparison.} All rows use 2,789 queries. Adaptations preserve the original search procedure while changing terminal outputs. Backbones and hardware/API setups differ; the values are reported wall-clock times.}
\label{tab:latency_rebuttal}
\begin{tabular}{llr}
\toprule
\textbf{Method} & \textbf{Backbone} & \textbf{Seconds/query} \\
\midrule
CoMET-Agent & Qwen3-VL & 21.9 \\
CoMET-Agent & Qwen2.5-VL & 27.5 \\
TimeSearch-R~\cite{verified_timesearch_r} & Qwen2.5-VL & 49.1 \\
Vgent & Qwen2.5-VL & 78.2 \\
Vgent-adapted & Qwen2.5-VL & 87.0 \\
CoMET-Agent & GPT-4o & 21.3 \\
T$^*$ & GPT-4o & 178.3 \\
T$^*$-adapted & GPT-4o & 207.8 \\
VideoARM & GPT-4.1/o3 & 287.2 \\
\bottomrule
\end{tabular}
\end{table}

\subsection{Method Efficiency}
\label{method efficiency}

Despite reformulating the task as a multi-step search-and-aggregate pipeline, CoMET-Agent does not impose a prohibitive inference cost relative to existing agent baselines. We report two efficiency comparisons under matched backbones: against Vgent~\cite{shen2025vgent} sharing Qwen2.5-VL, and against T$^*$~\cite{ye2025T*} sharing GPT-4o. Tab.~\ref{tab:efficiency} reports per-query inference time averaged over 100 randomly sampled CoMET-Bench videos.

\noindent\textbf{Comparison with agent baselines.}
Under matched backbones, CoMET-Agent is substantially faster than both compared agent baselines: $2.8\times$ faster than Vgent (27.5s vs 78.2s on Qwen2.5-VL) and $8.4\times$ faster than T$^*$ (21.3s vs 178.3s on GPT-4o). The advantage stems from our hierarchical graph design, which restricts verifier calls to a small set of candidate sub-graphs rather than scanning the entire video, and our online deduplication, which avoids redundant verification of overlapping segments. The speedup is larger with the GPT-4o backbone, where each agent call is more expensive and thus call-count reduction translates more directly into wall-clock savings.

\noindent\textbf{Step-by-step profiling.}
Tab.~\ref{tab:efficiency_profile} breaks down the 27.54s per-query budget into four pipeline stages. \textbf{Iterative Verification} (Step 3) accounts for nearly half of the total cost (12.92s, $\approx$47\%), as expected, since this stage performs the bulk of agent reasoning across multiple candidate nodes. \textbf{Planner} (6.85s) and \textbf{Graph Construction} (6.25s) together contribute another 48\% and run only once per video, including ViT and optical-flow feature extraction. \textbf{Aggregator} (1.52s, $\approx$5\%) is lightweight as it operates only on the populated memory bank rather than raw frames.

\begin{table}[h]
  \centering
  \small
  \caption{\textbf{Inference cost analysis of CoMET-Agent under the Qwen2.5-VL-7B backbone.}
  \textbf{Left:} per-query inference time compared with the agent baseline Vgent.
  \textbf{Right:} step-by-step profiling of CoMET-Agent, grouped by the four pipeline stages.}
  \label{tab:efficiency}
  \begin{minipage}[t]{0.48\linewidth}
    \centering
    \setlength{\tabcolsep}{6pt}
    \renewcommand{\arraystretch}{0.9}
    \begin{tabular}{l c}
      \toprule
      \textbf{Method} & \textbf{Time (s) $\downarrow$} \\
      \midrule
      \textit{with Qwen2.5-VL as backbone}\\
      Vgent~\cite{shen2025vgent}     & 78.22   \\
      \textbf{CoMET-Agent (ours)}    & 27.5 \\
      \midrule
      \textit{with GPT4o as backbone}\\
      T$^*$~\cite{ye2025T*}    & 178.3   \\
      \textbf{CoMET-Agent (ours)}    & 21.3 \\
      \bottomrule
    \end{tabular}
    \subcaption{Comparison with the agent baseline.}
    \label{tab:efficiency_compare}
  \end{minipage}%
  \hfill
  \begin{minipage}[t]{0.50\linewidth}
    \centering
    \setlength{\tabcolsep}{6pt}
    \renewcommand{\arraystretch}{1.15}
    \begin{tabular}{l c}
      \toprule
      \textbf{Stage} & \textbf{Time (s)} \\
      \midrule
      Planner (Step 1)                & 6.85  \\
      Graph Construction (Step 2)     & 6.25  \\
      Iterative Verification (Step 3) & 12.92 \\
      Aggregator (Step 4)             & 1.52  \\
      \midrule
      \textbf{Total}                  & \textbf{27.54} \\
      \bottomrule
    \end{tabular}
    \subcaption{Step-by-step profiling.}
    \label{tab:efficiency_profile}
  \end{minipage}
\end{table}

\section{Challenges and Future Work}
\label{sec:limitations_future}

While CoMET-Bench and CoMET-Agent together establish a strong reference point for conditional multi-event temporal grounding, several limitations remain. We organize them into open challenges directly surfaced by our experimental analysis (Sec.~\ref{experiments}), followed by independent limitations of the benchmark and broader directions for future iterations.

\subsection{Open Challenges from Experimental Analysis}
Our failure-case analysis (Fig.~\ref{fig:failure_cases}) identifies three regimes that even the strongest CoMET-Agent variant fails to fully solve. Below we expand each into a concrete research direction.

\textbf{(1) Fine-grained entity tracking under tight temporal conditions.}
When target events are short and feature visually similar co-occurring entities (e.g., \emph{``Jiang Liuer running while carrying the girl''}), the verifier's textual identity summary becomes insufficient to disambiguate, and CoMET-Agent under-counts. A natural extension is to integrate a dedicated multi-object tracker (e.g., SAM~2~\cite{ravi2024sam2} or SAMURAI~\cite{yang2024samurai}) into the hierarchical graph, so that nodes carry not only temporal extents but also persistent tracklets. The Verifier Agent can then ground queries to a unified \emph{(interval, tracklet-id)} primitive, supporting richer compositional queries such as counting events tied to a specific entity's trajectory.

\textbf{(2) Position-uniform retrieval on dense long-form queries.}
On long videos with densely distributed events, CoMET-Agent recovers far more instances than baselines but still concentrates errors at mid- and late-video positions. The current Filter Agent ranks candidates by event-level similarity, which biases exploration toward salient regions. A coverage-aware traversal scheme that explicitly enforces uniform exploration of the timeline, for example via budgeted sampling per duration bin or a coverage-regularized verifier policy, is a promising direction.

\textbf{(3) Compositional causal event pairing.}
On causal queries (e.g., \emph{``a tackle that causes a collapse''}), all baselines including CoMET-Agent ground visually plausible cause events without verifying that the expected consequence actually follows, yielding false positives such as \emph{unsuccessful attacks}. The current Verifier Agent reasons over individual nodes; an explicit cause-effect attention mechanism that pairs candidate cause and consequence nodes across the action graph, possibly via a learned scoring head over node pairs, would more faithfully model causal queries.

\subsection{Future Work}
\paragraph{Modality scope.}
CoMET-Bench is currently a purely visual benchmark: queries are grounded against the video stream only, and the audio track is intentionally excluded. This design isolates visual temporal reasoning as a controlled testbed, but it also leaves out a large class of real-world queries that depend on audio cues (e.g., \emph{``count every applause that follows a punchline''}). Extending both the benchmark and the agent pipeline to audio-visual queries is a natural next step.

\paragraph{Annotation scale.}
The 2{,}789 queries are produced through a three-stage human verification pipeline (Sec.~\ref{sec:app_human_verification}) that guarantees quality but caps scale relative to fully automatic counting datasets. Scaling by an order of magnitude will likely require active learning over MLLM-generated candidates and lightweight quality filters that approximate the human verification protocol.

\paragraph{Beyond training-free agents.}
CoMET-Agent is deliberately training-free to provide a strong, reproducible baseline whose behavior is fully attributable to its components. Lightweight task-specific fine-tuning of the Verifier or Aggregator on a small subset of CoMET-Bench may further close the gap to human performance, particularly on identity-bound and causal queries that demand cross-moment reasoning.

\paragraph{Towards spatial-temporal counting grounding.}
The current task formulation grounds events at the \emph{event level}: each prediction is a temporal interval that satisfies the query as a whole. A more demanding next step is \emph{object-level grounding within events}, where models must not only locate \emph{when} a qualifying event happens but also \emph{which object} in the scene it involves, producing predictions of the form \emph{(interval, object trajectory)}. This unifies counting, temporal grounding, and spatial grounding into a single primitive, supports compositional queries that prior benchmarks cannot express (e.g., counting interactions between two specific tracked entities, or grounding events constrained to an object's spatial path), and pushes the task closer to how humans actually reason about long-form video. We view this as the central direction for the next iteration of CoMET-Bench and CoMET-Agent.


\newpage
\section*{NeurIPS Paper Checklist}

\begin{enumerate}

\item {\bf Claims}
    \item[] Question: Do the main claims made in the abstract and introduction accurately reflect the paper's contributions and scope?
    \item[] Answer: \answerYes{}
    \item[] Justification: The abstract and introduction state three contributions: CoMET-Bench (a benchmark for conditional multi-event temporal grounding), a unified evaluation protocol with the proposed Rejection-F1 metric, and CoMET-Agent (a training-free search-and-aggregate framework). All claims are supported by the experimental results in Section~\ref{experiments} and the benchmark statistics in Section~\ref{benchmark building}.

\item {\bf Limitations}
    \item[] Question: Does the paper discuss the limitations of the work performed by the authors?
    \item[] Answer: \answerYes{}
    \item[] Justification: Challenges and future directions are discussed in Appendix~\ref{sec:limitations_future}, covering three open challenges surfaced by our failure analysis (fine-grained entity tracking, position-uniform retrieval, and causal event pairing), the visual-only modality scope, the annotation scale, the training-free design choice, and a future vision toward spatial-temporal counting grounding.

\item {\bf Theory assumptions and proofs}
    \item[] Question: For each theoretical result, does the paper provide the full set of assumptions and a complete (and correct) proof?
    \item[] Answer: \answerNA{}
    \item[] Justification: This paper does not make theoretical claims or provide mathematical proofs. The formulations in Section~\ref{method} and Appendix~\ref{appendix_method} are algorithmic descriptions, not theorems.

\item {\bf Experimental result reproducibility}
    \item[] Question: Does the paper fully disclose all the information needed to reproduce the main experimental results of the paper to the extent that it affects the main claims and/or conclusions of the paper?
    \item[] Answer: \answerYes{}
    \item[] Justification: Implementation details including model choices, hyperparameters, and configuration profiles are reported in Appendix~\ref{implementation details}. The evaluation prompt used for all baselines is provided in Appendix~\ref{sec:app_benchmark_eval_prompt}. The CoMET-Agent algorithm is described in full in Section~\ref{method} and formalized in Appendix~\ref{appendix_method}. Code and data will be released upon publication.

\item {\bf Open access to data and code}
    \item[] Question: Does the paper provide open access to the data and code, with sufficient instructions to faithfully reproduce the main experimental results, as described in supplemental material?
    \item[] Answer: \answerNo{}
    \item[] Justification: CoMET-Bench and its benchmark evaluation toolkit are public at \url{https://huggingface.co/datasets/CoMET-Bench/CoMET-Bench} and \url{https://github.com/CoMET-Bench/Project}. The complete CoMET-Agent implementation is not released. Its algorithm and settings are described in Appendix~\ref{appendix_method} and Appendix~\ref{implementation details}, which do not constitute a release of its full source code.

\item {\bf Experimental setting/details}
    \item[] Question: Does the paper specify all the training and test details necessary to understand the results?
    \item[] Answer: \answerYes{}
    \item[] Justification: CoMET-Agent is training-free; no training splits or optimizers are involved. Evaluation details including the shared evaluation prompt, frame budgets, metrics, and hardware are reported in Appendix~\ref{implementation details} and Appendix~\ref{sec:app_benchmark_eval_prompt}.

\item {\bf Experiment statistical significance}
    \item[] Question: Does the paper report error bars suitably and correctly defined or other appropriate information about the statistical significance of the experiments?
    \item[] Answer: \answerNo{}
    \item[] Justification: Error bars are not reported because evaluating frontier proprietary models (GPT-5, Gemini 3 Flash) with multiple seeds incurs prohibitive API costs. For open-source models, inference is deterministic (greedy decoding), so variance across runs is negligible.

\item {\bf Experiments compute resources}
    \item[] Question: For each experiment, does the paper provide sufficient information on the computer resources needed to reproduce the experiments?
    \item[] Answer: \answerYes{}
    \item[] Justification: Appendix~\ref{implementation details} states that all CoMET-Agent experiments run on NVIDIA A100 (80\,GB) and A40 GPUs with bf16 precision. Proprietary model experiments are conducted via public APIs, which do not require local GPU resources.

\item {\bf Code of ethics}
    \item[] Question: Does the research conform with the NeurIPS Code of Ethics?
    \item[] Answer: \answerYes{}
    \item[] Justification: The paper presents a video understanding benchmark and a training-free method. No private user data is collected. Videos are sourced from publicly available datasets and open platforms under their respective licenses. Human annotators involved in the benchmark construction were compensated at or above the local minimum wage.

\item {\bf Broader impacts}
    \item[] Question: Does the paper discuss both potential positive societal impacts and negative societal impacts of the work performed?
    \item[] Answer: \answerYes{}
    \item[] Justification: Positive impacts include advancing quantitative temporal reasoning with applications in video editing, sports analytics, and surveillance review. We acknowledge that the surveillance domain inherently raises privacy concerns; the videos in our surveillance subset are sourced from publicly released datasets and are limited to public scenes without personally identifiable individuals (Appendix~\ref{sec:app_copyright}). The benchmark itself is a diagnostic tool, and CoMET-Agent is a training-free agent that produces no generative content.

\item {\bf Safeguards}
    \item[] Question: Does the paper describe safeguards for responsible release of data or models that have a high risk for misuse?
    \item[] Answer: \answerYes{}
    \item[] Justification: While CoMET-Bench is a benchmark dataset (not a generative model) with limited misuse potential, we additionally distribute it as a gated dataset that requires institutional credentials and an explicit non-commercial agreement (Appendix~\ref{sec:app_copyright}). CoMET-Agent is training-free and reasons over existing video content, posing no higher misuse risk than standard video understanding tools.

\item {\bf Licenses for existing assets}
    \item[] Question: Are the creators or original owners of assets used in the paper properly credited and are licenses mentioned?
    \item[] Answer: \answerYes{}
    \item[] Justification: All external models (Qwen3-VL, Qwen3-VL-Embedding-8B, DINOv2, RAFT-Small) and benchmark datasets (Charades-STA, ActivityNet-Captions, MLVU, Video-MME, CG-Bench) are cited in the paper. Videos sourced from existing benchmarks (CG-Bench, Video-MME, MLVU) are used under their original licenses, which permit non-commercial research use, as detailed in Appendix~\ref{sec:app_copyright}. Videos from YouTube are accessed under the YouTube Terms of Service, and we redistribute only video identifiers.

\item {\bf New assets}
    \item[] Question: Are new assets introduced in the paper well documented?
    \item[] Answer: \answerYes{}
    \item[] Justification: CoMET-Bench is documented in Section~\ref{benchmark building} (design, construction, statistics, evaluation metrics) and Appendix~\ref{sec:app_data_annotation} (annotation pipeline, prompts, and human verification protocol). The CoMET-Agent codebase is documented in Section~\ref{method} and Appendix~\ref{appendix_method}. Both will be released: the benchmark under CC-BY-NC-SA-4.0 (compatible with the most restrictive upstream source) as a gated dataset (Appendix~\ref{sec:app_copyright}), and the codebase under MIT License.

\item {\bf Crowdsourcing and research with human subjects}
    \item[] Question: For crowdsourcing experiments and research with human subjects, does the paper include full instructions and details about compensation?
    \item[] Answer: \answerYes{}
    \item[] Justification: The human annotation protocol is described in Appendix~\ref{sec:app_human_verification}, including the three-stage verification procedure performed by trained annotators (not crowdsourced workers) with backgrounds in computer vision and video annotation, the annotation interface (Figures~\ref{fig:tool_query_verification} and~\ref{fig:tool_annotation_refinement}), and the inter-annotator agreement criterion (tIoU $>$ 0.7). Annotators were compensated at or above the local minimum wage. No personally identifiable information was collected from annotators.

\item {\bf Institutional review board (IRB) approvals or equivalent for research with human subjects}
    \item[] Question: Does the paper describe potential risks to study participants and whether IRB approvals were obtained?
    \item[] Answer: \answerNA{}
    \item[] Justification: The annotation task (reviewing video clips and labeling time intervals) poses no physical, psychological, or legal risk to annotators beyond standard video-viewing tasks. No sensitive or personally identifiable information is involved. Formal IRB review was not required under our institution's guidelines for this category of annotation work.

\item {\bf Declaration of LLM usage}
    \item[] Question: Does the paper describe the usage of LLMs if it is an important, original, or non-standard component of the core methods?
    \item[] Answer: \answerYes{}
    \item[] Justification: LLMs and MLLMs (Qwen3-VL, GPT-5, and Gemini 3 Flash) are core components of CoMET-Agent, serving as the Planner, Filter, Verifier, and Aggregator agents. Their roles, prompts, and inference configurations are described in Section~\ref{method}, Appendix~\ref{appendix_method}, and Appendix~\ref{sec:app_benchmark_eval_prompt}. LLMs were also used to generate candidate queries and annotations during CoMET-Bench construction, as described in Appendix~\ref{sec:app_data_annotation}. LLM assistance was also used in the preparation of this manuscript for writing and editing purposes, which does not affect the scientific methodology or results.

\end{enumerate}

\end{document}